%% file: main.tex
\documentclass[10pt,twocolumn,letterpaper]{article}

\input{configuration/config}

\usepackage[
style=ieee 
]{biblatex}

\usepackage[pagenumbers]{cvpr} 

\newcommand{\supplementarysection}{%
  \setcounter{section}{0}
  \setcounter{figure}{0}
  
  \let\oldthefigure\thefigure
  \renewcommand{\thefigure}{S\oldthefigure}
  
  \let\oldthetable\thetable
  \renewcommand{\thetable}{S\oldthetable}

  \let\oldthesection\thesection
  \renewcommand{\thesection}{S\oldthesection}
  
}

\title{Generative Proxemics: A Prior for 3D Social Interaction from Images}

\author{
   Lea M{\"u}ller\textsuperscript{1} \quad
   Vickie Ye\textsuperscript{2} \quad
  Georgios Pavlakos\textsuperscript{2} \quad
   Michael Black\textsuperscript{1} \quad
   Angjoo Kanazawa\textsuperscript{2}\\
   \textsuperscript{1}MPI for Intelligent Systems, T{\"u}bingen, Germany
   \quad
   \textsuperscript{2}UC Berkeley
   \\
   {\tt\small \{lea.mueller,black\}@tuebingen.mpg.de, \{mueller,vye,pavlakos,kanazawa\}@berkeley.edu}\\
}

\begin{document}

\begin{refsection}

\input{sections/02_Content}  
\clearpage
\input{sections/06_Disclaimer}
\printbibliography[heading=subbibintoc]
\end{refsection}

\begin{refsection}
\twocolumn[
    \centering 
    
    \section*{Generative Proxemics: A Prior for 3D Social Interaction from Images \\
    ** Supplementary Material **}
    
    \author{
       Lea M{\"u}ller\textsuperscript{1} \quad
       Vickie Ye\textsuperscript{2} \quad
       Georgios Pavlakos\textsuperscript{2} \quad
       Michael Black\textsuperscript{1} \quad
       Angjoo Kanazawa\textsuperscript{2}\\
       \textsuperscript{1}MPI for Intelligent Systems, T{\"u}bingen, Germany
       \quad
       \textsuperscript{2}UC Berkeley       \\
       {\tt\small \{lea.mueller,black\}@tuebingen.mpg.de, \{mueller,vye,pavlakos,kanazawa\}@berkeley.edu}\\
       \textcolor{white}{.}\\ 
    \textcolor{white}{.}\\
    }

]

\supplementarysection
\input{sections/04_Appendix}
\clearpage
\printbibliography[heading=subbibintoc]
\end{refsection}

\end{document}

%% file: configuration/config.tex
\input{configuration/01_Packages}
\input{configuration/02_Highlight}

\input{configuration/03_Style}
\input{configuration/05_Params}

\input{configuration/common_acronyms/bodymodels}
\input{configuration/common_acronyms/datasets}

\input{configuration/common_acronyms/math}

\input{configuration/common_acronyms/methods}

\input{configuration/common_acronyms/metrics}

\input{configuration/common_acronyms/common}

%% file: configuration/01_Packages.tex
\usepackage[accsupp]{axessibility}  
\usepackage[hyphens]{url}
\usepackage[dvipsnames]{xcolor}  
\usepackage{colortbl}
\usepackage{graphicx}
\usepackage{amsmath}
\usepackage{amssymb}
\usepackage{booktabs}
\usepackage{multirow}
\usepackage{cuted}
\usepackage{bm}
\usepackage{xspace}
\usepackage{caption}
\usepackage{balance}
\usepackage{pifont}
\usepackage[normalem]{ulem}
\usepackage{arydshln}
\usepackage{lipsum}
\usepackage{longtable}
\usepackage[utf8]{inputenc}
\usepackage{rotating}
\usepackage{csquotes}
\usepackage{epigraph} 

\usepackage[accsupp]{axessibility}  
\usepackage{colortbl}
\usepackage{graphicx}
\usepackage{amsmath}
\usepackage{amssymb}
\usepackage{booktabs}
\usepackage{multirow}
\usepackage{cuted}
\usepackage{bm}
\usepackage{xspace}
\usepackage{caption}
\usepackage{balance}
\usepackage{pifont}
\usepackage[normalem]{ulem}
\usepackage{arydshln}
\usepackage{lipsum}
\usepackage{times}
\usepackage{epsfig}
\usepackage{graphicx}
\usepackage{amsmath}
\usepackage{amssymb}
\usepackage{tabularx}
\usepackage[hyphens]{url}
\usepackage[dvipsnames]{xcolor}  
\definecolor{cvprblue}{rgb}{0.21,0.49,0.74}
\usepackage{listings}
\usepackage[pagebackref=false,breaklinks=true,colorlinks,bookmarks=false,citecolor=cvprblue]{hyperref}

%% file: configuration/02_Highlight.tex
\newcommand{\methodCOLOR}{black}



%% file: configuration/03_Style.tex
\newcommand{\colorRef}[1]{\textcolor{black}{#1}} 
\usepackage[capitalize]{cleveref}
\crefname{figure}{\colorRef{Fig.}}{\colorRef{Figs.}}
\Crefname{figure}{\colorRef{Figure}}{\colorRef{Figures}}
\crefname{section}{\colorRef{Sec.}}{\colorRef{Secs.}}
\Crefname{section}{\colorRef{Section}}{\colorRef{Sections}}
\Crefname{table}{\colorRef{Table}}{\colorRef{Tables}}
\crefname{table}{\colorRef{Tab.}}{\colorRef{Tabs.}}

\definecolor{codegreen}{rgb}{0,0.6,0}
\definecolor{codegray}{rgb}{0.5,0.5,0.5}
\definecolor{codepurple}{rgb}{0.58,0,0.82}
\definecolor{backcolour}{rgb}{0.95,0.95,0.92}

\lstdefinestyle{mystyle}{
    commentstyle=\color{codegreen},
    keywordstyle=\color{magenta},
    numberstyle=\tiny\color{codegray},
    stringstyle=\color{codepurple},
    basicstyle=\ttfamily\footnotesize,
    breakatwhitespace=false,         
    captionpos=b,                    
    keepspaces=true,                 
    numbers=left,                    
    numbersep=2pt,                  
    showspaces=false,                
    showstringspaces=false,
    showtabs=false,                  
    tabsize=4,
    morekeywords={with},
    otherkeywords={with},
}

%% file: configuration/05_Params.tex
\newcommand{\mesh}{M}

\newcommand{\vertexvec}{V}

\newcommand{\region}{r}

\newcommand{\contactmapbinary}{\mathcal{C}^D}

\newcommand{\pose}{\bm{\theta}}
\newcommand{\shape}{\bm{\beta}}
\newcommand{\orientation}{\bm{\phi}}
\newcommand{\orient}{\orientation}
\newcommand{\expression}{\bm{\psi}}

\newcommand{\translation}{\bm{\gamma}}
\newcommand{\trans}{\translation}

\newcommand{\energy}{L} 

\newcommand{\pone}{X^a}
\newcommand{\ptwo}{X^b}
\newcommand{\mone}{\mesh^a}
\newcommand{\mtwo}{\mesh^b}
\newcommand{\orienta}{\orient^{a}}
\newcommand{\orientb}{\orient^{b}}
\newcommand{\posea}{\pose^{a}}
\newcommand{\poseb}{\pose^{b}}
\newcommand{\shapea}{\shape^{a}}
\newcommand{\shapeb}{\shape^{b}}
\newcommand{\transa}{\trans^{a}}
\newcommand{\transb}{\trans^{b}}

\newcommand{\nverticessmplx}{10,475\xspace}

\newcommand{\nregionsflickrcithreeds}{75\xspace} 



%% file: configuration/common_acronyms/bodymodels.tex
\newcommand{\smplx}{\mbox{SMPL-X}\xspace}

\newcommand{\smplxa}{\mbox{SMPL-XA}\xspace}

\newcommand{\smpl}{\mbox{SMPL}\xspace}

\newcommand{\smil}{\mbox{SMIL}\xspace}
\newcommand{\smilx}{\mbox{SMIL-X}\xspace}

%% file: configuration/common_acronyms/datasets.tex
\newcommand{\flickrcithreeds}{\mbox{FlickrCI3D} Signatures\xspace}

\newcommand{\chithreed}{\mbox{CHI3D}\xspace}

\newcommand{\hifourd}{\mbox{Hi4D}\xspace}

%% file: configuration/common_acronyms/math.tex
\newcommand{\norm}[1]{\left\lVert#1\right\rVert}


%% file: configuration/common_acronyms/methods.tex
\newcommand{\bev}{\mbox{BEV}\xspace}

\newcommand{\openpose}{\mbox{OpenPose}\xspace}

\newcommand{\vposer}{\mbox{VPoser}\xspace}

\newcommand{\buddi}{\mbox{BUDDI}\xspace}

\newcommand{\vitpose}{\mbox{ViTPose}\xspace}

\newcommand{\methodname}{\mbox{{\color{\methodCOLOR}\mbox{BUDDI}}}\xspace}
\newcommand{\mn}{\methodname}

\newcommand{\modelName}{BUDDI\xspace} 
\newcommand{\modelNameLong}{BUDdies DIffusion Model\xspace} 



%% file: configuration/common_acronyms/metrics.tex
\DeclareSymbolFont{matha}{OML}{txmi}{m}{it}
\DeclareMathSymbol{\varv}{\mathord}{matha}{118}

\newcommand{\mpjpePA}{\mbox{PA-MPJPE}\xspace}

\newcommand{\pampjpe}{\mpjpePA}

%% file: configuration/common_acronyms/common.tex
\newcommand{\supmat}{Sup.~Mat.\/\xspace}

\newcommand{\twod}{2D\xspace}
\newcommand{\threed}{3D\xspace}

\newcommand{\sixd}{6D\xspace}
\newcommand{\rgb}{\mbox{RGB}\xspace}

\newcommand{\mocap}{\mbox{MoCap}\xspace}

\newcommand{\flickr}{\mbox{Flickr}\xspace}
\newcommand{\pgt}{\mbox{pseudo-ground truth}\xspace}

%% file: sections/02_Content.tex
\input{sections/03_BUDDI/figtex/teaser}

\begin{abstract}
\vspace{-1em}
Social interaction is a fundamental aspect of human behavior and communication. The way individuals position themselves in relation to others, also known as \textit{proxemics}, conveys  social cues and affects the dynamics of social interaction. Reconstructing such interaction from images presents challenges because of mutual occlusion and the limited availability of large training datasets. To address this, we present a novel approach that learns a prior over the 3D proxemics two people in close social interaction and demonstrate its use for single-view \threed reconstruction. 
We start by creating \threed training data of interacting people using image datasets with contact annotations. We then model the proxemics using a novel denoising diffusion model called \textit{BUDDI} that learns the joint distribution over the poses of two people in close social interaction. 
Sampling from our {\em generative proxemics} model produces realistic 3D human interactions, {which we validate through a perceptual study.} We use BUDDI in reconstructing two people in close proximity from a single image without any contact annotation via an optimization approach that uses the diffusion model as a prior. Our approach recovers accurate and plausible 3D social interactions from noisy initial estimates, outperforming state-of-the-art methods. 
Our code, data, and model are availableat our project website at: \url{muelea.github.io/buddi}.
\end{abstract}

\section{Introduction} 
\label{sec:introduction}

Humans are social creatures, and physical interaction plays a crucial role in our daily lives, shaping our relationships.
For example, research in behavioral science has shown that a slight touch between two people can cause a more friendly behaviour towards the touch-giver and lead to increased tips in restaurants \cite{crusco1984midas}. However, capturing and modeling scenarios of physical social interaction in three dimensions is a challenging task that requires a deep understanding of the intricate interplay between body poses, shape, and proximity. These interactions are hard to model by hand and best learned from data. 

In this work, we present the first approach that learns a generative model for 3D social proxemics and demonstrate its use as data-driven prior during an optimization routine. The diffusion model is trained using 3D human poses and shapes reconstructed from a large-scale image collection~\cite{fieraru2020three} using contact annotation, as well as using motion-capture (MoCap) data~\cite{fieraru2020three,yin2023hi4d}. The resulting model is able to generate the 3D pose and shape parameters of pairs of interacting people. When trained on bodies recovered from images, the model learns interactions depicted in photographs, such as people standing close together, playing sports, hugging, \etc, see Figure~\ref{fig:teaser_buddi}. We further demonstrate the effectiveness of the learned prior by applying it to the challenging task of \threed human pose and shape reconstruction from a single image containing people engaged in social interaction.

Specifically, we propose \modelName: a ``\modelNameLong''. Diffusion models are established methods for image generation and are often used to model \threed human motion. In this work we use them to model \threed social proxemics. The majority of state-of-the-art diffusion-based methods for 3D human mesh generation operate on 3D joint locations \cite{tevet2022human}. This representation lacks information about the human body surface, which, intuitively, is important for reasoning about interpersonal contact. Our approach, in contrast, operates on the parameters of two parametric human body models, which represent the surfaces of two people closely interacting. After training, our model is able to generate samples of plausible pairs of \threed bodies in social interaction from pure noise. The model can also be conditioned in the output of a human pose and shape regressor. In this conditional case, the model effectively takes the noisy output and generates similar poses but with realistic social interaction.

\looseness=-1 We then demonstrate how exploit \modelName's knowledge of human proxemics to guide \threed mesh reconstruction of people in a close social interaction from a single image. To this end, we introduce a novel optimization-based approach, which uses \modelName as a data-driven prior. 
{We initialize our optimization routine with samples from \mn, conditioned to the output of a state-of-the-art multi-person human mesh regressor \cite{sun2022putting}}. We then optimize over \smplx pose, shape, and translation parameters to match detected \twod joint locations. We incorporate guidance from the diffusion model using a loss inspired by the Score-Distillation loss from the 3D object creation literature \cite{poole2022dreamfusion}: In each optimization step, \modelName refines the current estimate towards a more plausible social interaction conditioned on the initial predictions. The refined pose, shape and translation serve as prior in the overall objective function. 

Our contributions include (1) presenting the first generative model of a pair of \threed people in close social interaction and (2) a novel approach for reconstructing \threed human meshes from images without relying on ground-truth contact annotations. 
We perform extensive experiments with \modelName to evaluate its performance on the FlickrCI3D Signatures dataset \cite{fieraru2020three} as well as CHI3D, and the recent Hi4D dataset \cite{yin2023hi4d} and find that it outperforms the state of the art as well as strong baselines. We also evaluate the unconditional samples from the diffusion model in a perceptual study, where people find our samples more realistic 44.4\% when compared over real samples, where 50\% is the upperbound where they do not see any difference.
Importantly, we find that our optimization approach significantly improves the results of \cite{sun2022putting} both quantitatively and qualitatively.
This work opens up a new avenue of research on digital human synthesis, laying the foundation for a deeper understaning of human social behavior derived from image data. Our data, code, and model will be available for research.

\input{sections/03_BUDDI/figtex/method.tex}

\section{Related Work} 
\label{sec:relatedwork}

\textbf{Generating \threed humans.} There has been recent interest in  generating \threed humans, in different contexts. Several methods automatically populate static \threed scenes with \threed humans \cite{hassan2021populating,zhang2020generating,zhang2020place}, while more recent methods generate both body and hand poses to interact with \threed objects \cite{Taheri:CVPR:2022,wu2022saga,tendulkar2023flex}. Other work generates human motions conditioned on different inputs such as audio \cite{li2021ai,tseng2022edge} or text \cite{petrovich2021action,petrovich2022temos,tevet2022human}. Concurrent work proposes text-to-\threed diffusion-based approaches to generate motion of two interacting humans \cite{liang2023intergen,shafir2023human}. Neither method predicts the full body surface, but rather they synthesizes either \threed joint locations or \smpl pose parameters for the average body.
These methods are not used as priors for reconstructing interacting people from images.

To model 3D human proxemics probabilistically, we employ diffusion models, which achieve impressive performance on image generation tasks \cite{dhariwal2021diffusion,ho2020denoising,rombach2022high,saharia2022photorealistic}. They have recently been adopted in \threed human motion generation scenarios: MDM~\cite{tevet2022human} generates plausible motions conditioned on text input; PhysDiff~\cite{yuan2023physdiff} incorporates physical constraints in the diffusion process to generate physically plausible motions; and EDGE~\cite{tseng2022edge} uses a transformer-based diffusion model for dance generation. Related work~\cite{chen2023executing,dabral2022mofusion,ma2022pretrained} has investigated different modalities for the conditioning, \eg, audio, text, or action classes. EgoEgo~\cite{li2023ego} generates plausible full-body motions conditioned on the head motion. SceneDiffuser~\cite{huang2023diffusion} focuses on the scene-conditioned setting. We also rely on techniques from the diffusion literature, but consider the unique setting where two people are in close interaction and leverage this for single-image \threed reconstruction. 

\textbf{Multi-person \threed human mesh estimation.} An extensive line of work focuses on reconstructing the \threed human pose and shape of a single person from images using optimization~\cite{bogo2016keep,guan_iccv_scape_2009,lassner2017unite,Pavlakos2019_smplifyx,rempe2021humor,tiwari22posendf,xu2020ghum} or regression approaches~\cite{arnab2019exploiting,guler2019holopose,joo2021exemplar,Kanazawa2018_hmr,Kolotouros2019_spin,mueller2021tuch,omran2018neural,xu2019denserac,zanfir2021neural,zhang2021pymaf}. Capitalizing on these techniques, recent approaches focus explicitly on reconstructing multiple people jointly from a single image. Zanfir~\etal~\cite{zanfir2018monocular} propose an optimization solution, while Jiang~\etal~\cite{jiang2020multiperson} and Sun~\etal~\cite{sun2021monocular} rely on deep networks to regress the pose and shape for all people in the image. BEV~\cite{sun2022putting} extends ROMP \cite{sun2021monocular} to reason about the depth of people in a virtual birds-eye-view while taking age/height into account. 
We use BEV~\cite{sun2022putting} as an initialization for our optimization method, and demonstrate how our  learned 3D social proxemics prior improves the estimation of close human-human interactions.

The above methods do not address contact between people. To do so, Fieraru~\etal~\cite{fieraru2020three} introduce the first datasets with ground-truth labels for the body regions in contact between humans. Labels are collected using MoCap (\chithreed) or human annotators (\flickrcithreeds). They propose an optimization approach that requires the ground-truth contact map to reconstruct people in close proximity at test time. 
More recently, REMIPS \cite{fieraru2021remips} introduces a transformer-based method that regresses the \threed pose of multiple people. REMIPS is trained using the above datasets while taking into account contact and interpenetration. In this work, we take a very different approach by learning and exploiting a 3D generative proxemics prior. We use the ground-truth contact maps to generate pseudo-ground truth \threed human fits from which we learn the diffusion model; once this is learned, we show that it can be used as a prior to recover plausible bodies in close proximity from images without explicit knowledge of contact maps. 

\textbf{Data-driven priors in optimization.} Optimization-based methods for \threed human pose and shape estimation, like SMPLify~\cite{bogo2016keep},  are versatile and allow different data-driven prior terms to be incorporated in the objective function. Different methods have been used to learn pose priors including GMMs \cite{bogo2016keep}, VAEs \cite{Pavlakos2019_smplifyx}, neural distance fields \cite{tiwari22posendf}, and normalizing flows \cite{zanfir2020weakly}. ProHMR~\cite{kolotouros2021probabilistic} learns a pose prior conditioned on image pixels. HuMoR~\cite{rempe2021humor} incorporates a data-driven motion prior in the iterative optimization. POSA~\cite{hassan2021populating} learns a prior for human-scene interaction from PROX data~\cite{hassan2019resolving} and uses it in their optimization. In contrast to these methods, we use a diffusion model to capture the joint distribution over \smplx parameters for two people interacting and show that we can both sample from the model and use during optimization to improve the pose estimates of interacting people.

\section{Method} 
\label{sec:method}

We introduce \buddi, a generative model of two people in close social interaction.
Because of the complexity and multimodality of the data, we turn to denoising diffusion probabilistic models \cite{ho2020denoising} to address this task. In \cref{sec:diffusion_model}, we describe the basics of diffusion, and the parameterization we employ to model people in contact. In addition to sampling new body meshes from our model, our generative model can serve as a prior for reconstructing 3D humans from images. In \cref{sec:buddi_prior}, we describe an optimization procedure that incorporates BUDDI as a prior to recover two \smplx meshes from observed 2D keypoints.

For all of the following, we use the \smplx~\cite{Pavlakos2019_smplifyx} body model to represent the human bodies. 
\smplx is a differentiable function that maps pose, $\pose \in \mathbb{R}^{21\times3}$, shape, $\shape \in \mathbb{R}^{10}$, and expression, $\expression\in \mathbb{R}^{10}$ parameters to a mesh consisting of $N_{v}=\nverticessmplx$ vertices $\vertexvec \in \mathbb{R}^{N_{v} \times 3}$. We place the generated meshes in the world by rotating and translating them by $\orient \in \mathbb{R}^{3}$ and$\trans\in\mathbb{R}^{3}$. We denote person $a$'s parameters as $\pone = [\orienta, \posea, \shapea, \transa]$ and $\ptwo = [\orientb, \poseb, \shapeb, \transb]$. For simplicity, we refer to both people when no index is specified, \eg, $X$ refers $\pone$ and $\ptwo$
 
\subsection{Diffusion Model for 3D Proxemics}
\label{sec:diffusion_model}

  \newcommand{\tf}[1]{\mathbf{#1}}
Denoising diffusion models are latent variable generative models that learn to transform random noise into the desired data distribution $p_{\text{data}}$
through a forward and reverse process. 
The forward inference process is a Markov chain over $T$ steps given by transitions
$q(\tf{x}_{t+1} | \tf{x}_t)$, 
which gradually adds Gaussian noise to clean samples $\tf{x}_0$ from the data distribution according to a fixed variance schedule $\sigma_t$.

The reverse process $q(\tf{x}_{t-1} | \tf{x}_t)$ then gradually denoises noisy samples back into the data distribution. 
The reverse process transitions follow a Gaussian distribution when conditioned on $x_0$, but must be inferred during the generative process.
Following \cite{ramesh2022hierarchical}, we train
a neural network $D$ that predicts a sample
$\tf{\hat x}_0 = D(\tf{x}_t;t)$ 
from a noisy sample $\tf{x}_t$ given the noise level $t$.For the task of reconstructing humans from images, when we have initial estimates of the \smplx parameters of two humans,
we condition the denoising network $D$ on $\tf{c}_H$, the predicted \smplx parameters of the two humans by a regressor such as BEV~\cite{sun2022putting}.

We refer to the process of adding noise as \textit{diffusion} and the process of removing the noise via $D$ as \textit{denoising}.
Specifically, we diffuse a ground-truth sample $\tf{x}_0$ 
by uniformly sampling a noise level~$t$ with \mbox{$\epsilon_t \sim \mathcal{N}(0, \tf{I})$} to obtain
the noisy sample \mbox{$\tf{x}_t = \sqrt{\sigma'_t} \tf{x}_0 + \sqrt{1 - \sigma'_t} \epsilon_t$} with $\sigma'_t = \prod_{i=1}^{t} (1 - \sigma_t).$

We then train $D$ to minimize
\begin{equation}
 \mathbb{E}_{\tf{x}_0 \sim p_{\text{data}}}    \mathbb{E}_{t\sim\mathcal{U}\{0, T\},\tf{x}_t \sim q(\cdot | \tf{x}_0)} ||D( \tf{x}_t; t, \tf{c}_H) - \tf{x}_0 ||,
\end{equation}
where we set $\tf{c}_H = \varnothing$ for 20\% of conditional model training, and all of unconditional model training.

\paragraph{Architecture.}
Because we aim to model close contact between people, we choose a model state space that can express the full surface of the human body.
Specifically, in contrast to prior work in human motion diffusion that operate only on joint angles and locations~\cite{tevet2022human,yuan2023physdiff},
we directly operate on the full \smplx parameters of the two people.
A sample $\tf{x}$ thus corresponds to the concatenation of two bodies:
$$\tf x =[\pone, \ptwo] = [\orienta, \posea, \shapea, \transa, \orientb, \poseb, \shapeb, \transb] \text{.}$$

We denoise a sample $\tf{x}_t$ with a transformer encoder block on tokenized parameters.
Specifically, each parameter of each person is tokenized into 152-dimensional latent vectors with per-parameter and per-person embedding layers. We tokenize the noise level $t$ similarly with a noise embedding.
When conditioning is available, i.e. SMPL-X estimates for person $a$ and $b$,
we similarly tokenize the parameters to be used as additional tokens.
We pass the available tokens into the transformer encoder, and similarly decode the output tokens with per-token embeddings.
We illustrate the denoiser architecture in Figure~\ref{fig:diffusion}.

\textbf{Losses.}
We employ standard human pose and shape regularization losses.
We write our training objective as
\begin{equation}
    L_D = L_{\theta} + L_{\beta} + L_{\trans} + L_{v2v} \text{, }
\end{equation}
where $L_\theta$, $L_\beta$, $L_{\trans}$ denote squared L2-losses on respective body model parameters, and $L_{v2v}$ denotes a squared L2 loss on model vertices.
We use \sixd rotation representations~\cite{zhou_2019_cvpr} for global orientation and pose,
and model the relative translation between $a$ and $b$.
We show generated samples from our unconditional model in \cref{fig:samples}.

\input{sections/03_BUDDI/figtex/samples}

\subsection{Optimization with the Proxemics Prior} 
\label{sec:buddi_prior}
Reconstructing 3D human meshes from a single image is an extremely under-constrained problem, and priors over human pose and shape are crucial in an optimization based framework 
for recovering plausible meshes~\cite{bogo2016keep,Pavlakos2019_smplifyx,zanfir2018monocular}.
Our problem involves people in close contact, which requires correctly placing the meshes in context with each other, which has only been done when given ground truth contact annotations at test time~\cite{fieraru2020three}.
We remove the need for ground-truth contact maps
by using our generative model as a prior during reconstruction with a score distillation approach~\cite{poole2022dreamfusion,sjc}.

During inference, we observe detected \twod keypoints ${\tilde J_{2D}}$ and initial body model parameter estimates $\tf{c}_H$ from a regressor~\cite{sun2022putting}.
We then optimize the body parameters of two people to minimize
\begin{equation}
    L_{\text{Optimization w. \mn}} = L_{\text{fitting}} + L_{\text{diffusion}}.
\end{equation}
$L_{\text{fitting}}$ ensures that the solution stays close to the image evidence,
while $L_\text{diffusion}$ is a data-driven prior using our conditional diffusion model.
We treat this prior as similar to those used for \threed pose in previous works such as GMM~\cite{bogo2016keep} and V-Poser~\cite{Pavlakos2019_smplifyx}, but for 3D proxemics. 
We illustrate the optimization procedure in \cref{fig:diffusion} right.

We initialize our optimization by generating a sample $\tf{\tilde{x}}$ from the conditional model.
We sample with DDIM sampling with 100 evenly spaced steps.
We then use the data fitting loss:
\begin{equation}
    \begin{split}
      L_{\text{fitting}} = & \lambda_{J} \energy_{J} + \lambda_{\tilde{\theta}} \energy_{\tilde{\theta}} + \lambda_{P} \energy_{P},
    \end{split}
\end{equation}
where $\energy_{J}$ denotes  \twod re-projection error between the reprojected \threed joints of the current estimate and the detected \twod keypoints, and
$\energy_{\tilde{\theta}}$ is a prior for the solution to be close to the denoised initialization. 

$\energy_{P}$ denotes an interpenetration loss between two people that pushes inside vertices to the surface,
which we compute using winding numbers between low-resolution \smplx meshes of the current estimates:
\begin{equation}
    \begin{aligned}
         \energy_{P} = \sum_{v \in\vertexvec^{a}_{I}} \min_{u \in \vertexvec^{b}} \norm{v - u}^2 +
        \sum_{v \in \vertexvec^{b}_{I}} \min_{u \in \vertexvec^{a}} \norm{v - u}^2 \text{,} 
    \end{aligned}
\end{equation}
where $\vertexvec^a_I$ denotes vertices of 
$\mone$ intersecting the low-resolution mesh of $\mtwo$;
and vice versa for $\vertexvec^b_{I}$.

To use the prior on human interaction into account, we use the learned denoising model $D$ from \mn and perform a single \textit{diffuse-denoise} step, {with a noise level at $t=10$}, 
on the current estimate.
The denoised estimate, $\hat{\tf{x}}_0 = D(\tf{x}_t; t, \tf{c}_H)$, 
regularizes the current estimate via
\begin{equation}
    \begin{split}
        {L_\text{diffusion}} & = || D(\tf{x}_t; t, \tf{c}_H) - \tf{x} || \text{,}
    \end{split}
\end{equation}
where $\tf{x}_t = \sqrt{\sigma'_t} \tf{x}_{\texttt{no-grad}} +  \sqrt{1 - \sigma'_t} \epsilon_{t}$ denotes the diffused body model parameters of the current estimate,
and $\tf{x}_{\texttt{no-grad}}$ denotes the current estimate with detached gradients.
$\tf{\hat{x}}_0$, and encourages $\tf{x}$ to be close to $\hat{\tf{x}}_0$.
In practice, we penalize the decoded parameters of $\tf{x}$ and $\hat{\tf{x}}_0$ directly as
\begin{equation}
\begin{split}
  L_\text{diffusion} =
  &\lambda_{\hat{\orient}} || \hat{\tf{\orient}_0} - \tf{\orient} || + \lambda_{\hat{\pose}} || \hat{\tf{\pose}_0} - \tf{\pose} ||\\
  &+ \lambda_{\hat{\shape}} || \hat{\tf{\shape}_0} - \tf{\shape} || + \lambda_{\hat{\trans}} || \hat{\tf{\trans}_0} - \tf{\trans} || \text{.}
\end{split}
\end{equation}
Intuitively, this loss uses the learned denoiser~$D$ to take a step from the current estimate towards the data distribution of two people in close proximity, conditioned on the  regressor prediction.

\section{Implementation Details} 
\label{sec:training_buddi}

\textbf{Training Data.} There are few datasets containing \threed ground truth of humans in close social interaction \cite{yin2023hi4d,fieraru2020three}. Such datasets are usually captured in lab environments, consequently they are small and do not contain the variety of interactions between humans ``in the wild,'' \eg when playing sports or taking social pictures. To address this lack of data, we create \textbf{\flickr Fits}, \ie \smplx fits for \flickr images portraying humans in contact scenarios. For this, we use \flickrcithreeds \cite{fieraru2020three}, a dataset of images showing interacting humans collected from \flickr with discrete human-human contact annotations. Specifically, the \smplx body surface is divided into $R=\nregionsflickrcithreeds$ regions such that each region, $\region$, roughly covers a similar area. For a given photo, the human annotators assign a binary label indicating contact between a region on one person and a region on the other.
For two meshes, $\mone$ and $\mtwo$, the annotation can be represented as a binary contact map $\contactmapbinary \in \{0,1\}^{R \times R}$, where
\begin{equation}
  \contactmapbinary_{ij}=\left\{
  \begin{array}{@{}ll@{}}
    1, & \text{if $r_i$ of $\mone$ is in contact with $r_j$ of $\mtwo$} \\
    0, & \text{otherwise.}
  \end{array}\right.
\end{equation}

We use these ground-truth contact maps in an optimization routine for fitting two people to detected keypoints, similar to \cref{sec:buddi_prior} but replace the diffusion model prior with standard image fitting priors. Please see the \supmat for a description of this process and see \cref{fig:psgt} for qualitative examples. The dataset contains 10,631/1,139 train/test images, with one image containing multiple contact annotations. Note that we only use images containing matching \bev outputs, \twod keypoints, and contact labels.

\input{sections/03_BUDDI/figtex/psgt}

We also augment our training data with available \mocap data, which is considerably smaller than those obtained from image fits:
\textbf{\chithreed}~\cite{fieraru2020three} contains 3/2 pairs of training/test subjects performing 127 sequences of two-person interactions like hugs or kicks with ground-truth \smplx bodies. One frame per sequence has contact map annotations. We use the contact frame of the sequences from two subject pairs, resulting in 247 mesh pairs for training, and the third pair for evaluation. \textbf{\hifourd}~\cite{yin2023hi4d} contains sequences of 20 pairs of people interacting with each other. The interactions include actions like hugging, dancing, and fighting. We randomly split the data into 14/3/3 pairs for train/val/test and use every fifth frame after the firest contact between the two subjects, resulting in about 1K mesh pairs for training. The body representation format in Hi4D is \smpl, which we transfer to \smplx using the \smplx code repository \cite{Pavlakos2019_smplifyx}. Please see the \supmat for more details of the datasets. Note that while we use \smplx model, \mn is not trained on hands because none of these datasets contain hand poses.

\textbf{\mn Training.} {\mn} is trained with meshes from \flickrcithreeds Fits, \chithreed, and \hifourd. We use 60\%~\flickr, 20\%~\chithreed, and 20\%~\hifourd data distribution per batch with batch size 512. The transformer backbone has six layers and eight heads; we use 10\% dropout and randomly shuffle the order of people during training. To train \mn, we randomly sample noise levels $t$ up to 1000 using a cosine noise schedule~\cite{nichol2021improved}. We use the Adam optimizer \cite{kingma2014adam} with learning rate $10^{-4}$.
We train two versions of \mn, an unconditional model for generation, and the conditional version for reconstruction. For the conditional model, we use all camera views of the \mocap datasets, \ie 4/8 cameras for CHI3D/Hi4D. The unconditional model is trained on \threed \mocap fits in the world coordinate system.
To sample new poses, we use DDIM sampling starting at noise levels $t=1000$ in steps of 10.

\textbf{Optimization Details.} During optimization, we experiment with different noise levels, between $10$ and $100$, and find that $t=10$ does not disturb the inputs too much, but enough for $D$ to generate new configurations. We use \bev \cite{sun2022putting} estimated as conditioning and detected \twod keypoints from OpenPose \cite{OpenPose_PAMI} and ViTPose \cite{xu2022vitpose}. 
Please see \supmat for more details.

\section{Experiments} 
\label{sec:experiments_buddi}

\noindent \textbf{Baselines.} We compare our reconstruction method with \bev \cite{sun2022putting}, which is also used as an input to our conditional model. Since there is no other available work that reasons about people in close social interaction, we experiment with simple but effective baselines. We train the transformer model of \mn to directly predict \smplx parameters of people in contact from \bev input, essentially a deterministic, single-step ablation of \mn. We also evaluate the direct conditional denoised output of \bev by \mn without any optimization.
As another baseline, we propose an optimization routine that replaces $L_{\text{diffusion}}$ with a simple heuristic that takes the minimal distances between two meshes predicted by \bev and minimizes their distance during optimization along with the other energy terms. Finally, to compare the generation ability we train a VAE which we also use during the optimization routine in a similar manner to \vposer \cite{Pavlakos2019_smplifyx} but for two people by optimizing the VAE latent space instead of \smplx parameters. We refer to these models as \textit{Transformer}, \textit{\mn (gen.)}, \textit{Contact Heuristic}, and \textit{VAE}, respectively. All baselines are trained on the same datasets as \mn with the same sampling strategies. More details about our baseline models are provided in the \supmat.

\noindent \textbf{Metrics.} We use standard evaluation metrics from the human pose and shape estimation literature. We also report the joint \pampjpe computed by performing Procrustes alignment of both people together. In addition to per-person metrics, this captures the relative orientation and translation of the two people. 
Since our method directly estimates 3D humans 
we propose a new metric similar to PCK~\cite{yang2012articulated} from the 2D pose literature called \textbf{PCC}, the percentage of correct contact points with respect to a radius $r$. Specifically, given two meshes, $\mone/\mtwo$ and a contact map $\contactmapbinary$ we compute the pairwise vertex-to-vertex Euclidean distances $d_\text{eucl}(\contactmapbinary)$ between annotated contact regions and consider the pair to be correct when $\min(d_\text{eucl}(\contactmapbinary)) < r$. 

\input{sections/03_BUDDI/figtex/bigfigure}

\subsection{Unconditional Generation}

\input{sections/03_BUDDI/tables/flickrci3d_val}

\input{sections/03_BUDDI/tables/hi4d}
We qualitatively evaluate \mn~by showing samples generated from the model in \cref{fig:teaser_buddi} and~\ref{fig:samples}.
Our approach is able to generate people in close proximity including embraces, handshakes, having a conversation, sitting side by side, and, in general, plausibly interacting with each other. Since it is trained on Internet image collections, it also learns to generate people posing for photographs or playing sports. 

We further run a perceptual study to evaluate the realism of the generated social interactions against other methods. In a forced choice study, we compare our generated samples with samples from the real data distribution according to the 60/20/20 per-batch ratio for \flickr/\chithreed/\hifourd used during training. We also compare \mn against generations from the VAE and a non-parametric random  
baseline that samples meshes from the pseudo-ground truth after centering the two people. We do a forced choice comparison between \mn and these there other methods, asking workers on Amazon Mechanical Turk to choose the sample that shows a more realistic close social interaction. We use 256 samples per method.  We collect ratings for 768 pairwise comparisons. In this study, \mn was chosen over random in 71.23\% of the comparisons, over the VAE in 60.17\%, and over the training data in 44.4\%.
Note that 50\% is the upper bound for such forced choice comparisons,
in which participants cannot tell the difference between real and generated samples.

For a quantitative evaluation, we compute the FID score between samples from \mn and samples from the \textit{VAE} on concatenated \smplx parameters. We sample 8K examples per method and from our training data following the dataset ratio per batch. \mn has a lower FID score (1.6) compared to the \textit{VAE} (3.3).

\subsection{Fitting with BUDDI}
\input{sections/03_BUDDI/tables/chi3d}
We show qualitative results in Fig.~\ref{fig:bigfig} comparing \mn against \bev and the Contact Heuristic.
Our approach is able to generate various types of human interactions with plausible contact and depth placement.
It is also able to capture close interaction between a child and a parent.
Although the Contact Heuristic (center) is able to move two people closer together, which helps with image alignment, upon close observation it is not able to capture the subtle interaction between people that happens during intimate interaction. \mn's estimates are more realistic and better capture the subtle details of interaction.
We provide additional qualitative examples in the \supmat

We further report the percentage of correct contact (PCC) with respect to the ground truth contact map on the \flickrcithreeds test set in \Cref{tab:flickr_test_result_mpjpe}.
The table also shows the pose reconstruction accuracy against our \flickr Fits. All metrics show improvement over \bev, in particular the joint \pampjpe.
Non-optimization methods, \ie \textit{Transformer} and \textit{\mn (gen.)}, are able to predict plausible contacts, with similar PCC accuracy to \mn,
but struggle to reconstruct the data with a worse joint PA-MPJPE.
The \textit{Heuristic}, in contrast, achieves a lower reconstruction error, but worse PCC.
Our approach which leverages the learned prior during optimization can recover both the relative positions and contacts between the two people.

We further evaluate our model against ground truth \mocap data in \Cref{tab:chi3d_test_result_mpjpe} and \Cref{tab:hi4d_test_result_mpjpe}.
Optimization with \mn consistently improves the two-person reconstruction error over \bev and other baselines.
When evaluated per action, the strongest improvements over \bev come from complex close social interactions like hugging or kissing, at 58mm and 54mm absolute improvement over BUDDI  respectively. 
The \textit{Heuristic} baseline achieves a low PA-MPJPE reconstruction error on all three datasets but is not sufficient to recover the joint poses.
\textit{Transformer} and \textit{\mn (gen.)} have lower joint PA-MPJPE errors than \bev and the \textit{Heuristic}, but worse per-person reconstruction errors.
The \textit{VAE} results suggest that directly operating in the latent space of a generative model is challenging and not sufficient to accurately recover close social interactions.
\mn, in contrast, is able to model a wide variety of poses, as supported by the numerical results. 

\section{Conclusion} 
\label{sec:conclusion}
We propose \mn, a diffusion model for close human-human interaction. We train \mn from 3D fits obtained from a large-scale dataset of images with ground truth contact annotations as well as a small set of available mocap data. \mn enables unconditional sampling of people in close social interaction. More importantly, we also demonstrate how \mn can be used as an effective prior for single-view 3D reconstruction of pairs of people in close proximity. 

Our core contribution of a generative proxemics prior provides the foundation for future work on modeling and capturing human interaction.
For example, future work could iteratively apply our method to new images and use the reconstructed examples to further improve the generative prior. Additionally, conditioning modalities can be explored, \eg, conditioning on pixel features, on text, or on action labels.
Future work could explore more fine-grained interactions that include finger pose and even facial expressions. Finally, these insights could be also extended to 3D motion capture and also interactions that involve more than two humans.

%% file: sections/03_BUDDI/figtex/teaser.tex
\twocolumn[{
    \renewcommand\twocolumn[1][]{#1}
    \maketitle
    \centering
    \includegraphics[width=1.00 \linewidth,clip,trim=0cm 14.5cm 0cm 0cm]{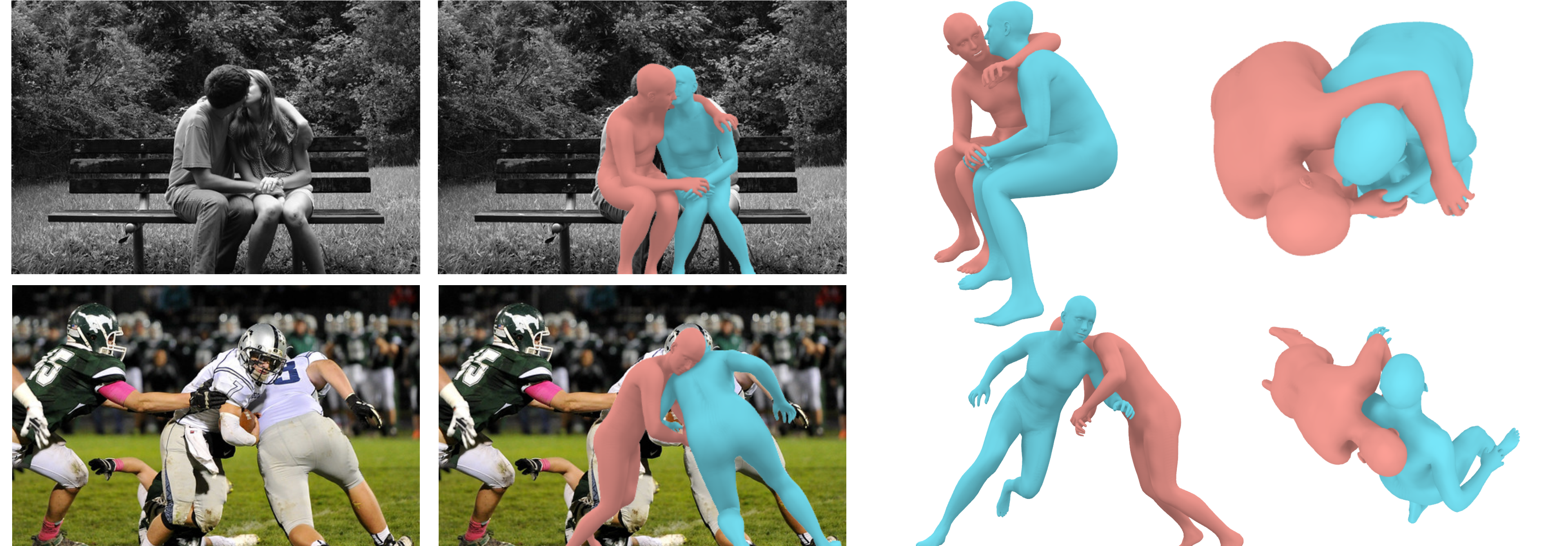}
    \captionof{figure}{\textbf{Generative Proxemics.} We propose a diffusion model that learns a 3D generative model of two people in close social interaction. We show how the model can be used to generated samples or as a social prior in the downstream task of reconstructing a pair of people in close proximity from images without any user annotation at test time. Shown here on the left are input test images, our predicted 3D bodies on the right.
    }
    \label{fig:teaser_buddi} 
    \vspace{1em}
}]

%% file: sections/03_BUDDI/figtex/method.tex
\begin{figure*}[t]
    \centering
\includegraphics[width=\linewidth]{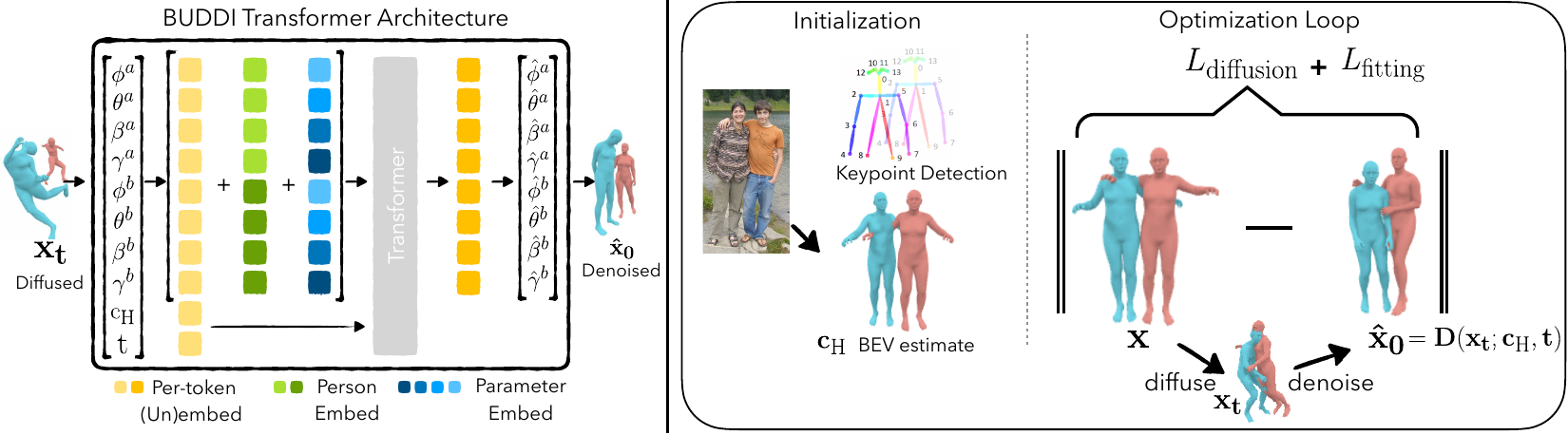}
\caption[Illustration of the architecture of BUDDI and our optimization routine.]{\textbf{BUDDI: BUddies DIffusion model.} On the left, we illustrate the architecture of BUDDI, our diffusion model for modeling 3D social proxemics between two people in close interaction. The diffusion process is applied directly on \smplx body parameters. To condition \mn on estimated body model parameters, $c_\text{H}$, we concatenate the parameters along the token dimension. On the right, we illustrate the optimization method with BUDDI as prior. Our optimization takes detected keypoints \cite{xu2022vitpose,OpenPose_PAMI} and an initial regressor estimate \cite{sun2022putting} as input. Given the regressor estimate, we sample from \mn to obtain $\tilde x$ which we use to initialize the optimization routine. In each optimization iteration, we take a single \textit{diffuse-denoise} step on the current estimate using the learned denoiser model $D$ conditioned on the initial BEV estimate. Our losses encourage the current estimate to be close to the refined meshes ($L_{\text{diffusion}}$) and to the initial estimate and detected keypoints ($L_{\text{fitting}}$).}
\label{fig:diffusion}
\end{figure*}

%% file: sections/03_BUDDI/figtex/samples.tex
\begin{figure*}[t]
    \centering
\includegraphics[width=\textwidth,clip,trim=0cm 12.5cm 0cm 0cm]{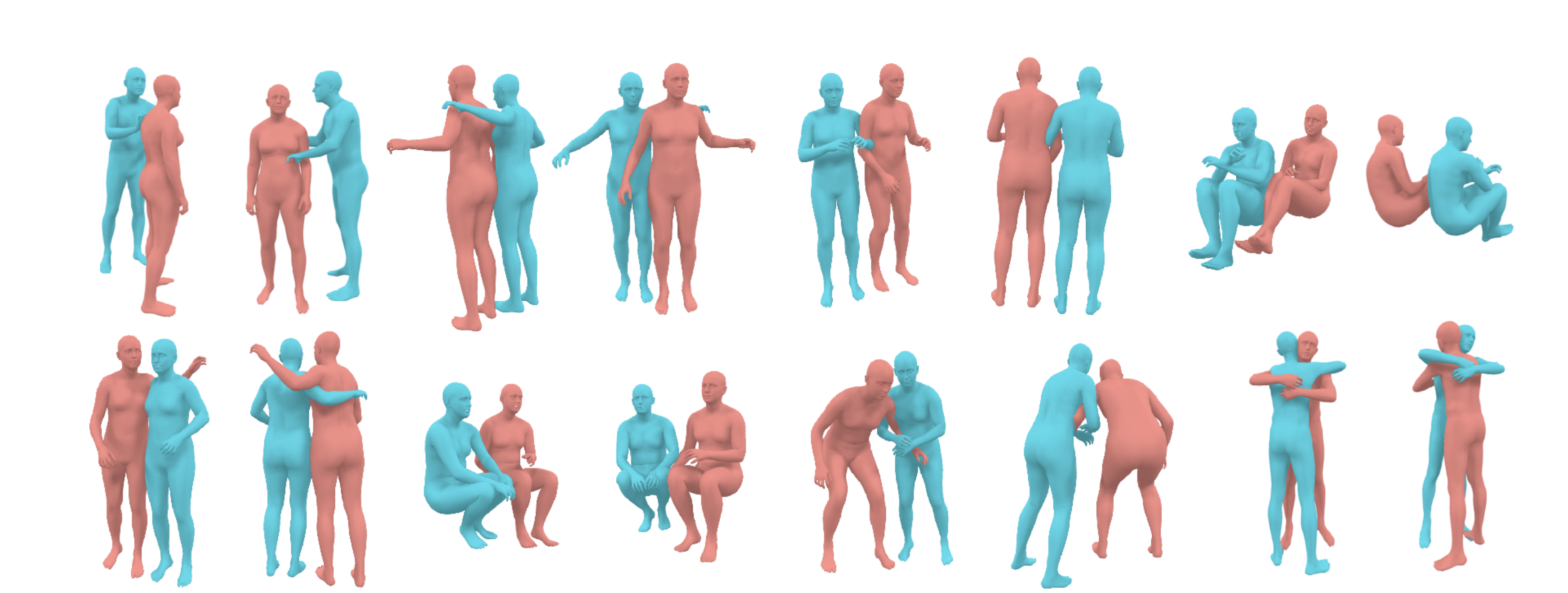}
\caption[Unconditional generation of meshes using BUDDI]{\textbf{Generative Proxemics: Samples from \buddi.} All samples are unconditionally generated from pure noise using the trained diffusion model. We select several representative examples and show two views per sample. 
These samples reveal that \buddi has learned the distribution of people in close contact including embracing each other, playing sports, sitting side by side, and taking photographs.}
\label{fig:samples}
\end{figure*}

%% file: sections/03_BUDDI/figtex/psgt.tex
\begin{figure}[t]
    \centering
\includegraphics[width=\linewidth,clip,trim=1cm 2.5cm 1cm 2.5cm]{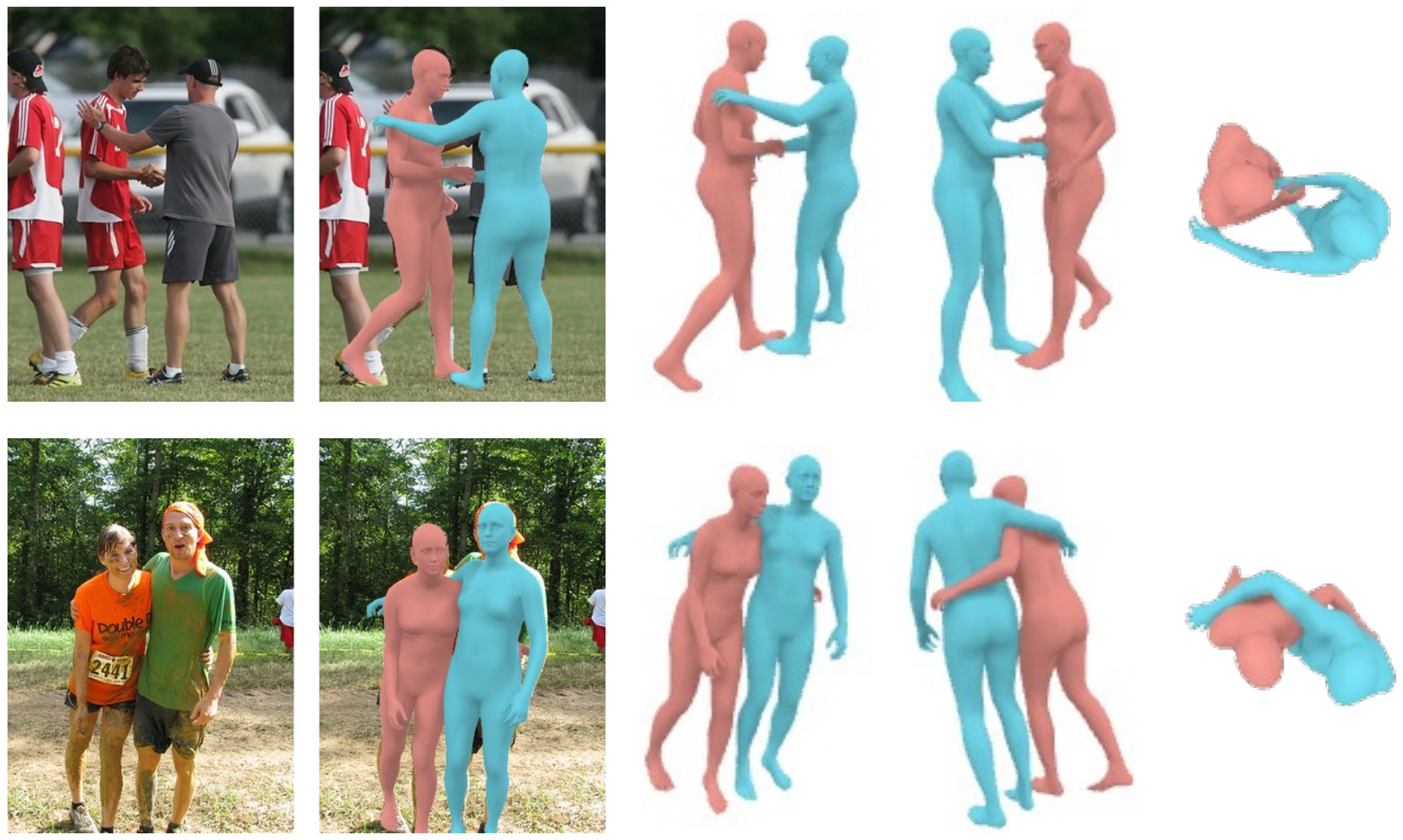}
\caption[Flickr Fits]{\textbf{Flickr Fits.} We visualize the output of the optimization process that reconstructs two people in close proximity using ground-truth contact maps, shown from four different views. We use these 3D fits as training data for BUDDI. Please see \supmat for more results.} \vspace{-1em}
\label{fig:psgt}
\end{figure}

%% file: sections/03_BUDDI/figtex/bigfigure.tex
\begin{figure*}[t]
    \centering
\includegraphics[width=\textwidth,clip,trim=0cm 0cm 0cm 0cm]{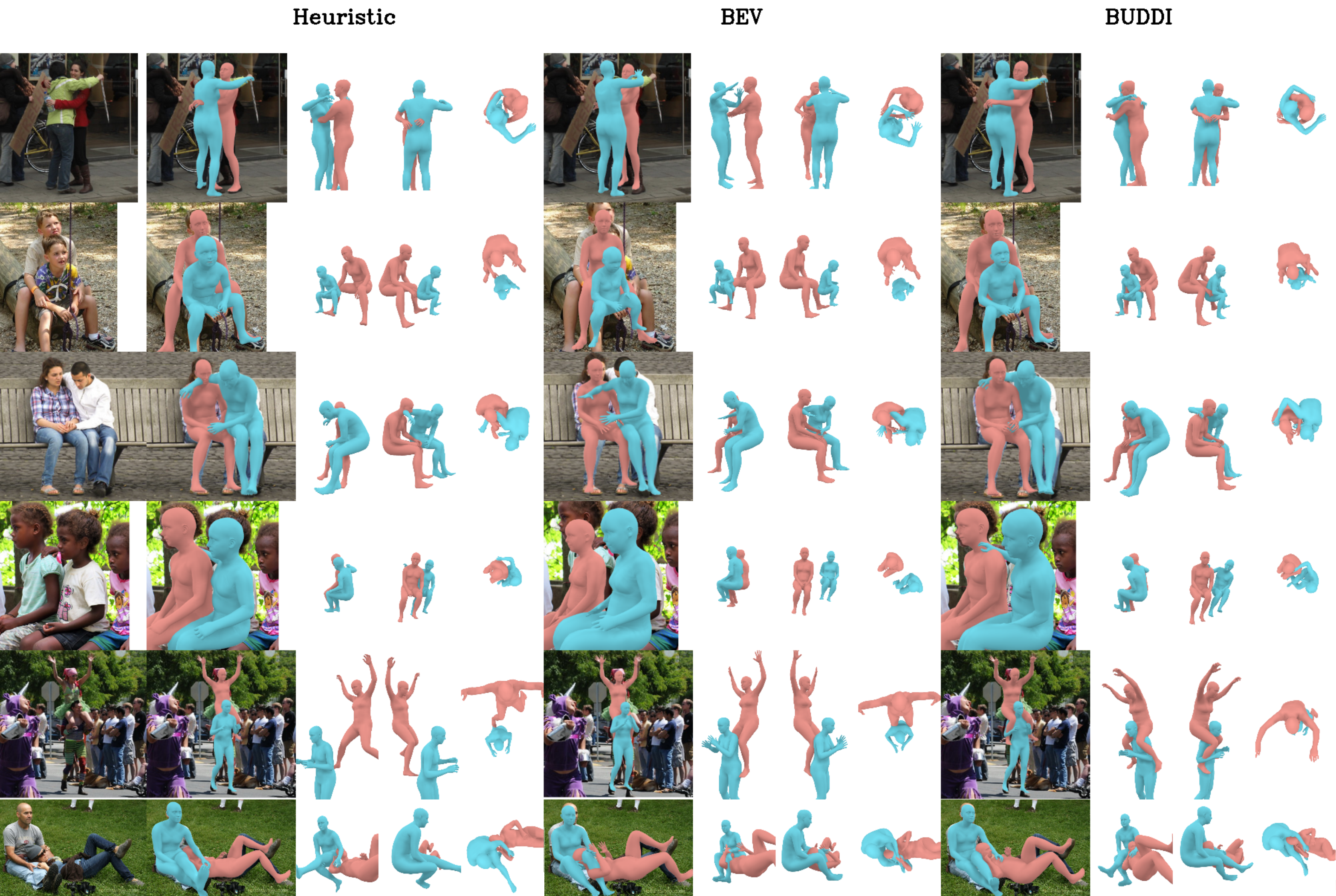}
\caption{\textbf{Automatic reconstruction of people in close social interaction.} 
We show qualitative results from a) BEV, b) contact heuristics, which takes the BEV output and encourages the closest parts to be in contact, and c) our method, which optimizes the BEV estimates against the image evidence with the BUDDI prior. Our approach recovers a plausible reconstruction with subtle details.}
\label{fig:bigfig}
\end{figure*}

%% file: sections/03_BUDDI/tables/flickrci3d_val.tex
\addtolength{\tabcolsep}{-2pt}
\begin{table}[t]
\scriptsize
\begin{center}
\begin{tabular}{lcccccccccccccc}
        \toprule[0.4mm]
        &    \multirow{2}{*}{\shortstack{JOINT $\downarrow$ \\ PA-MPJPE}}  &  \multicolumn{5}{c}{PCC at radius $\uparrow$}  \\
        &       &   5 & 10 & 15 & 20 & 25 \\ 
        \midrule
\bev           & 106     & -       & -       & -       & -       & -        \\
Transformer    & 86      & 14      & 40      & 60      & \textbf{73}      & \textbf{82}       \\
\mn (gen.)     & 92      & 15      & 39      & 58      & 71      & 80       \\
\midrule
Heuristic      & 68      & 14      & 34      & 49      & 61      & 70       \\
VAE            & 205     & 6       & 15      & 23      & 30      & 36       \\
\mn            & \textbf{66}      & \textbf{19}      & \textbf{44}      & \textbf{62}      & \textbf{73}      & 81       \\
        \bottomrule[0.4mm]
\end{tabular}
\end{center}
\vspace{-0.8em}
\caption[Evaluation of 3D Pose on \flickrcithreeds]{\textbf{3D Pose Evaluation on \flickrcithreeds.} We evaluate methods against the \flickr fits using their joint (two-person) PA-MPJPE expressed in mm. We also evaluate the percentage of correct contact points (PCC) for radius \textit{r} mm.
}
\label{tab:flickr_test_result_mpjpe}
\end{table}
\addtolength{\tabcolsep}{3pt}

%% file: sections/03_BUDDI/tables/hi4d.tex
\begin{table*}
\scriptsize 
    \centering
        \begin{tabular}{lcc|cccccccccccc} 
        \toprule[0.4mm]
     &    PER PERSON $\downarrow$ & JOINT $\downarrow$ & \multicolumn{11}{c}{JOINT PA-MPJPE $\downarrow$} \\
     & PA-MPJPE & PA-MPJPE & backhug    & basketball & cheers     & dance      & fight      & highfive   & hug        & kiss       & pose       & sidehug    & talk                   \\
    \midrule
    BEV                                                                                          & 78         / 84         & 136        & 200        & 126        & 109        & 135        & 121        & 106        & 163        & 139        & 142        & 131        & 118       \\
    Heuristic  & \textbf{67}  / \textbf{71}        & 121        & 168        & \textbf{83}         & 94         & 131        & \textbf{94}         & \textbf{68}         & 159        & 159        & 118        & 113        & 109       \\
    BUDDI (F, C)     & 70         / 77         & 115        & 200       & 94        & 92        & 128        & 108        & 100        & 133        & 114       & 104      & 107       & 91       \\
    \midrule
    Transformer      & 79         / 85         & 120        & 161        & 141        & 103         & 138        & 123        & 128        & 117        & 106         & 120        & 105        & 100    \\
    BUDDI (gen.)       & 82         / 90         & 117        & 152        & 139        & 120        & 137        & 130        & 96         & 101        & 97         & 115        & 102        & 101       \\
    VAE            & 80        / 82         & 138        & 175        & 133        & 114        & 141        & 119        & 87          & 176        & 162        & 135        & 140        & 113       \\
    BUDDI         & 70         / 76         & \textbf{98}         & \textbf{127}        & 95         & \textbf{92}         & \textbf{113}        & 109        & 72         & \textbf{105}        & \textbf{85}         & \textbf{88}         & \textbf{96}         & \textbf{81}  \\
        \bottomrule[0.4mm]
        \end{tabular}
    \caption[Evaluation of BUDDI on Hi4D]{\textbf{Evaluation of BUDDI on Hi4D.} We compare the output of \mn to the proposed baseline methods on the Hi4D challenge. The first block shows methods that do not use \hifourd data during training or are optimization based without access to priors trained on \hifourd. \mn (F,C) in particular, is our model \mn trained on \flickr and CHI3D data only. All errors are reported in mm for 3D Joints.}
    \label{tab:hi4d_test_result_mpjpe}
\end{table*}  

%% file: sections/03_BUDDI/tables/chi3d.tex
\begin{table}
\centering
\resizebox{.7\columnwidth}{!}{

    \begin{tabular}{lccc}
        \toprule
                    & PER PERSON  $\downarrow$           & JOINT $\downarrow$   \\
                    & PA-MPJPE               & PA-MPJPE \\
        \midrule
    BEV             & 50       \ 52          & 96      \\
    Transformer     & 54       \ 56          & 105     \\
    BUDDI (gen.)     & 53       \ 53          & 80      \\
        \midrule
    Heuristic       & 49     \textbf{46}     & 105     \\
    VAE             & 54       \ 54          & 103     \\
    BUDDI           & \textbf{48} \ 47       & \textbf{68}      \\
        \bottomrule
    \end{tabular}}
    \caption[Evaluation of BUDDI on \chithreed]{\textbf{Quantitative Evaluation on \chithreed.} We compare the output of our model to the baselines on \chithreed (pair s03). All errors reported in mm for 3D Joints.}
    \label{tab:chi3d_test_result_mpjpe}
\end{table}  

%% file: sections/06_Disclaimer.tex
\textbf{Acknowledgements.}
We thank our colleagues for their feedback, in particular, we thank Aleksander Holynski, Ethan Weber, and Frederik Warburg for their discussions about diffusion and the SDS loss, Jathushan Rajasegaran, Karttikeya Mangalam and Nikos Athanasiou for their discussion about transformers, and Alpar Cseke, Taylor McConnell and Tsvetelina Alexiadis for running the user study. Lea Müller is supported by the International Max Planck Research School for Intelligent Systems (IMPRS-IS). 

\textbf{Disclosure.}
This project was funded in part by NSF:CNS-2235013, Society of Hellman Fellows, and BAIR/BDD sponsors. MJB has received research gift funds from Adobe, Intel, Nvidia, Meta/Facebook, and Amazon.  MJB has financial interests in Amazon, Datagen Technologies, and Meshcapade GmbH.  While MJB is a consultant for Meshcapade, his research in this project was performed solely at, and funded solely by, the Max Planck Society.

%% file: sections/04_Appendix.tex
\section{Creating Flicker Fits}

To train BUDDI, we need 3D poses for two people interacting in close proximity.
We create this training data by fitting \smplx to \flickrcithreeds \cite{fieraru2020three} with an optimization method that takes ground-truth contact annotations into account. \flickrcithreeds is a publicly available datasets consisting of images collected from \flickr with \threed contact annotations. This is a complex task that requires data preprocessing steps. First, images in \flickr may contain children and infants who are not supported by \smplx \cite{Pavlakos2019_smplifyx}. Consequently, we follow previous work \cite{patel2021agora} and merge \smplx with \smilx \cite{hesse2018learning} to represent a range of body shapes from children to adults. 
Second, we observe that keypoints detected by ViTPose \cite{xu2022vitpose} are more accurate than those detected by OpenPose \cite{OpenPose_PAMI}, especially when people are occluding each other, except for the feet, which are often not detected. Therefore we merge OpenPose and ViTPose for our optimization method. Third, our method takes multiple modalities as input (keypoints and SMPL bodies estimated by \bev \cite{sun2022putting}, and ground truth contact maps for \flickr images) and uses these in estimating high-quality SMPL bodies.  
Fourth, the detections of \bev are in \smpl format, while the ground-truth contact maps for \flickr are provided for the \smplx template mesh. 
Since these two body models are not compatible, \ie their pose and especially their shape space is different, we create ``approximate'' \bev estimates in \smplx format by using the \smpl pose as if it was \smplx while properly converting the body shape from \smpl to \smplx.

\subsection{Preprocessing} 
\label{sec:appxbuddipreprocessing}

In the next section, we describe the preprocessing steps performed to create \flickr fits. We use the same preprocessing for our optimization with \mn.

\paragraph{Including children.} Since \smpl~\cite{SMPL:2015} only models adult body shapes, most human pose and shape regressors do not consider child body shapes explicitly. However, we found that \flickrcithreeds includes images of children (roughly 10\% of the images). Following the SMPLA~\cite{patel2021agora} convention, \bev also estimates a scale parameter~$s$, which is used to interpolate between \smpl \cite{SMPL:2015} (adult model) and \smil \cite{hesse2018learning} (infant model) for the template meshes and shape blend shapes. A scale value of $s=0.0$ is equivalent to \smpl only, a scale value of $s=1.0$ is equivalent to \smil only, and all the values in between  model intermediate stages. 
To extend this from \smpl to \smplx, we use the scale parameter estimated by \bev to interpolate between the \smplx and the \smilx template and shape blend shapes in \smplx topology. We visually found that this interpolation works well for $s\leq 0.8$, so we exclude pairs where the detected scale is $s>0.8$ for one of the interaction partners.
In practice, we concatenate the interpolation and body shape parameters such that $\shape \in \mathbb{R}^{11}$. We refer to this model as \smplxa.

\paragraph{Matching input detections.} 
As input, we have the estimated 3D bodies from \bev \cite{sun2022putting} and we have a dataset of ground-truth human-human contacts. The bodies in these two data sources are not in correspondence. To generate the \flickr Fits, we must first automatically put them in correspondence so that we can optimize the \bev bodies by exploiting the ground truth contact information.

In particular, we have (1) detected meshes from \bev, (2) \twod keypoint detections from \vitpose \cite{xu2022vitpose}, and (3) ground-truth bounding boxes indicating the interacting pair of humans.
We observed that the ground-truth bounding boxes typically match with the bounding boxes surrounding  \openpose \cite{OpenPose_PAMI} keypoint detections.
As a result, we only need to correspond the \openpose detections with \vitpose detections and the \bev bodies.
Since we can reproject the 3D joints from BEV bodies to 2D keypoints, both correspondence problems require us to solve the assignment between sets of 2D keypoints.
To do this, we compute a keypoint-cost matrix taking the detection confidence scores into account.
We only consider keypoints with confidence score greater than $0.6$ (for \bev all keypoints have by default a score of $1.0$ due to the amodal prediction of the human body).
We make assignments in a greedy way, while also set a threshold ($0.008)$ to discard matches with large matching distance.

\paragraph{Merging keypoints.} Qualitatively, we found that \vitpose performs better than \openpose, particularly for people that are heavily occluded. Since \vitpose (unlike \openpose) does not detect keypoints on the feet, we can merge the \vitpose pose detections with feet keypoints detected by \openpose. We perform this extension only if the L2 distance between \vitpose and \openpose ankles is less than $5$ pixels. Additionally, since many images in \flickrcithreeds include people who are truncated below the waist, we often have missing or wrong keypoint detections for the lower body. Because of this, we use the projected \bev ankle joints, when the ankle keypoint detection confidence score is less than $0.2$. Finally, the original keypoint values $k_{\text{orig}}$ are normalized by the keypoint bounding box size via $k = k_{\text{orig}} / (\max(\text{bb}_{\text{height}}, \text{bb}_{\text{length}}) * 512)$. These steps give us a set of 2D keypoints that we use to generate the Flickr fits via an optimization method described below.

\paragraph{\smpl to \smplx body shape conversion.} Our method takes \bev estimates as input and optimizes them to fit the image evidence. Since \bev estimates meshes in \smpl topology and the ground-truth contact maps are provided in \smplx format, we transfer the \bev estimate to \smplx. Ideally, one would fit \smplx to \smpl via optimization. This process is time consuming and we found that it is sufficient to initialize the optimization routine by using the \smpl pose parameters with the \smplx body. 
For body shape, we  solve for the \smplx body shape using a simple least-squares optimization. The shaped vertices, $V_{\text{\smpl}}$ and $V_{\text{\smplx}}$, are obtained via
\begin{equation}
\begin{split}
    V_{\text{\smpl}} & = T_{\text{\smpl}} + D_{\text{\smpl}} \beta_{\text{\smpl}} \text{, and} \\
    V_{\text{\smplx}} & = T_{\text{\smplx}} + D_{\text{\smplx}} \beta_{\text{\smplx}},
\end{split}
\end{equation}
where $T_{\text{\smpl}}$ and $T_{\text{\smplx}}$ are the \smpl and \smplx template meshes, $D_{\text{\smpl}}$ and $D_{\text{\smplx}}$ the shape blend shapes, and $\beta_{\text{\smpl}}$ and $\beta_{\text{\smplx}}$ the shape parameters. Only $\beta_{\text{\smplx}}$ is unknown. Since the topology between \smpl and \smplx is different, we use a \smpl-to-\smplx vertex mapping $M \in \mathbb{R}^{10475 \times 6890}$, such that $D_{\text{\smplx}} = M D_{\text{\smpl}}$. Then we can directly solve for body shape, $\beta_{\text{\smplx}}$, in a least-squares manner:
$$\beta_{\text{\smplx}} = (D^{T}_\text{\smplx} D_\text{\smplx})^{-1} D^{T}_\text{\smplx} M D_\text{\smpl} \beta_\text{\smpl}.$$

\paragraph{Additional details.}
We use the first 10 shape components and keep the facial expression and finger pose fixed. Note that, although we use \smplx, we do not optimize hand pose due to the lack of 3D data of close human interaction with hands, as well as the lack of robust finger keypoint detectors for people in close proximity.
Extending this work to include detailed hand contact would be interesting future work.

\subsection{Optimization with ground-truth contact annotations}
We create \smplxa fits for \flickr images using ground-truth contact annotations. We use these fits to train our generative models, along with a small set of \mocap \threed poses. We also use them to evaluate the pose estimation error (JOINT PA-MPJPE) in Table 1 in the main manuscript.

Optimization-based methods for fitting \threed meshes to \rgb images usually rely on sparse signals, like \twod keypoints (ground-truth or detected), and priors for human pose and shape \cite{bogo2016keep,Pavlakos2019_smplifyx,zanfir2018monocular}. Only a few methods explicitly use self-~\cite{mueller2021tuch} or human-human~\cite{fieraru2020three} contact in their optimization.

Our optimization method takes as input the discrete human-human contact annotations and, for each person, detected \twod keypoints \cite{OpenPose_PAMI,xu2022vitpose}, and initial estimates for their pose, ${\tilde \pose}$, orientation, $\tilde \orient$, shape, ${\tilde \shape}$, and translation, ${\tilde \trans}$, which are provided by the output of \bev \cite{sun2022putting}. 

Given these inputs, we take a two-stage approach: In the first stage, we optimize pose, $\pose$, shape, $\shape$, and translation, $\trans$, encouraging contact between discretely annotated body regions, while allowing the bodies to intersect. In the second stage, we activate a new loss term to resolve human-human intersection. The output of the first stage is usually close to the final pose with only slight intersections, because of which we optimize only pose and translation and hold the body shape constant in stage two. The objective function is:
\begin{equation}
    \begin{split}
      L_{\text{Cmap-fitting}} = & \lambda_{J} \energy_{J} + \lambda_{\bar{\theta}} \energy_{\bar{\theta}} +  \lambda_{\theta} \energy_{\theta} + \\
      & \lambda_{\shape} \energy_{\shape} + \lambda_{P} \energy_{P} + \lambda_{\contactmapbinary} \energy_{\contactmapbinary} \text{,}
    \end{split}
\end{equation}
where $\energy_{J}$ denotes the \twod re-projection error, $\energy_{\bar{\theta}}$ is a prior on the initial pose, $ \energy_{\theta}$ is a
Gaussian Mixture Model pose prior \cite{bogo2016keep}, and $\energy_{\shape}$ an {L2-prior} that penalizes deviation from the \smplx mean shape. The discrete human-human contact loss, $\energy_{\contactmapbinary}$, minimizes the distance between vertices, $v/u$, assigned to regions, $r$, with annotated discrete human-human contact via:
\begin{equation}
    \energy_{\contactmapbinary} = \sum_{i,j} \contactmapbinary_{ij}
    \min_{v \in \region_i, u \in \region_j}{\norm{v - u}^2} \text{.}
\end{equation}
$\energy_{P}$ denotes an interpenetration loss, active in the second stage only, that pushes inside vertices to the surface. We use winding numbers to find intersecting vertices between two meshes, $\mone$ and $\mtwo$, and vice versa. This operation is usually slow and memory intensive, which is why we use low-resolution meshes of \smplx with only 1K vertices.
With $\vertexvec^a_I$ we denote vertices of 
$\mone$ intersecting the low-resolution mesh of $\mtwo$; 
$\vertexvec^b_{I}$ follows the same notation. The intersection loss term is defined as:
\begin{equation}
    \begin{aligned}
         \energy_{P} = \sum_{v \in\vertexvec^{a}_{I}} \min_{u \in \vertexvec^{b}} \norm{v - u}^2 +
        \sum_{v \in \vertexvec^{b}_{I}} \min_{u \in \vertexvec^{a}} \norm{v - u}^2 \text{.} 
    \end{aligned}
\end{equation}
We find functional weights, $\lambda$, for each term in the objective function (see \Cref{tab:training_weights}). The results of this fitting approach are illustrated in Figure~\ref{fig:psgt_supmat}. We use this optimization routine to reconstruct interacting people depicted in the \flickrcithreeds ~\cite{fieraru2020three}.

\input{sections/03_BUDDI/figtex/psgt_supmat}

\section{Diffusion model}
\label{sec:appxbuddi}
\paragraph{Transformer architecture.} To embed each body model parameter $x_{ij}$ of person $j \in \{1, 2\}$ and parameters $i \in \{\orient, \theta, \beta, \trans\}$ of size $d_i$ in the latent space dimension $d_l=152$, we use linear-SiLU-linear sequences:
$$f_{ij}(x_{ij}) = \text{SiLU}( x_{ij} A^{T}_{ij} + b_{ij}) B^{T}_{ij} + c_{ij},$$ 
where $A_{ij} \in \mathbb{R}^{d_l \times d_i}$, $b_{ij} \in \mathbb{R}^{d_l}$, $B_{ij} \in \mathbb{R}^{d_l \times d_l}$, and $c_{ij} \in \mathbb{R}^{d_l}$. After passing these parameters through the transformer, we again use a linear-SiLU-linear sequence to project them back into their original dimension $d_i$.

\begin{figure}
    \centering
    \includegraphics[width=\linewidth]{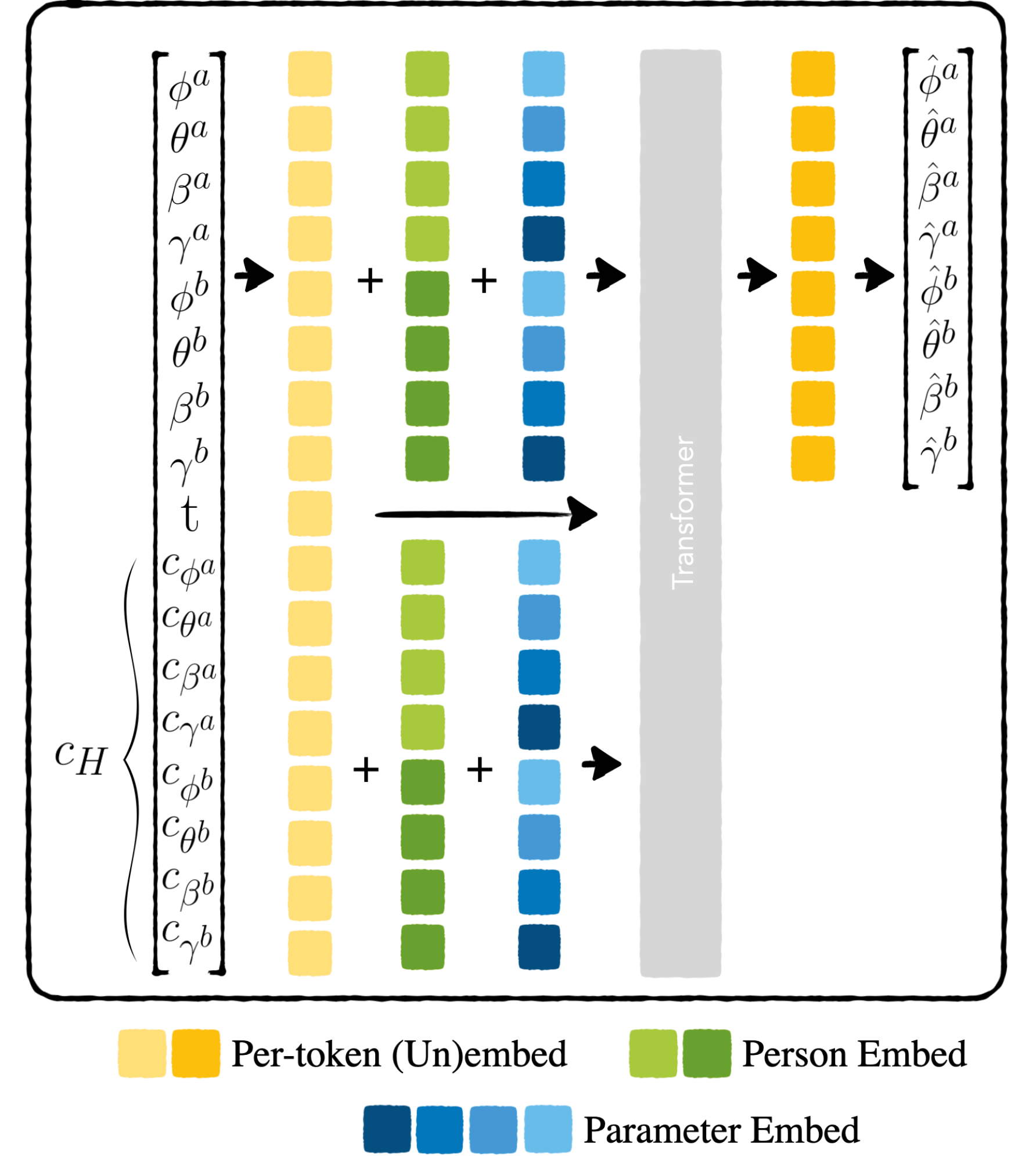}
    \caption{\textbf{Detailed architecture of \mn with conditioning.} When \mn is conditioned on model parameters, $c_\text{H}$, detected from \bev \cite{sun2022putting}, we concatenate the detected parameters (body global orientation, pose, shape, and translation for person a/b), with the input parameters along the token dimension and add per-person and per-parameter embedding vectors.}
    \label{fig:buddi_cond_detail}
\end{figure}

When BUDDI is trained with \bev \cite{sun2022putting} conditioning, we embed the conditioning in a similar fashion as the ground truth parameters, concatenate them along the token dimension, and add per-person and per-parameter embedding layers.  In \cref{fig:buddi_cond_detail}, we show the design of our conditional model. 

\section{Optimization}
Here, we provide additional information for the optimization routines, \ie optimization with contact map, VAE, heuristic, and BUDDI prior. In \Cref{tab:training_weights} we define the weights of each loss term. Every optimization runs for a maximum of 1000 iterations per stage, except optimization with \mn which we stop after 100 iterations. For termination, we use early stopping and we keep track of the loss value at the latest 10 iterations.
We use these values to fit a line with linear regression $f(x) = ax + b$ and terminate if $a < -1e-4$. We run each optimization for two stages. The second stage's reference poses, $\theta_0$, which are used in $\energy_{\tilde{\theta}}$, are taken to be the output / last pose of the first stage. 
We provide pseudo code in \cref{buddi_opti} showing the optimization routine with \mn used as prior.

\section{Training and Testing Datasets}
\subsection{Flickr Fits}
We split the Flickr \cite{fieraru2020three} training images into training and validation sets and use the provided test split for testing.
Fits can be noisy for example, when the assignment between contact annotations and keypoints is wrong or when keypoint detectors fail badly. To provide a reliable test set for \threed pose for images taken in the wild, we manually curate the Flickr Fits test set and detect 24 out of 1427 noisy fits. The final curated Flickr Test dataset contains 1403 interactions. We do not curate the training dataset.

\subsection{Hi4D}
Hi4D \cite{yin2023hi4d} is a \mocap dataset containing interaction between 20 pairs of people. Each pair performs about five interactions such as dancing, fighting, hugging, doing yoga, talking, etc. We split this dataset by subject pair into 14/3/3 for train/val/test. We use subjects [00, 01, 02, 09, 10, 13, 14, 17, 18, 21, 23, 27, 28, 37] for training, $[16, 19, 22]$ for validation, and [12, 15, 32] for testing. Since Hi4D was originally provided in \smpl format, we fit \smplx to the estimates via optimization using the code provided in the \smplx repository \cite{Pavlakos2019_smplifyx}. The dataset provides a start and end frame from/to which each sequence involves physical contact between two people. We use every 5th frame from the contact sequence for training and testing.

\subsection{CHI3D}
CHI3D \cite{fieraru2020three} is a \mocap dataset containing interactions between 3 pairs of people. Each pair performs eight interactions (grab, handshake, hit, holding hands, hug, kick, posing, and push) in various ways summing up to a total of about 120 sequences per subject pair. We use subjects [02, 04] for training and leave [03] for evaluation. Each sequence has a single frame with contact labels. We use this frame from each sequence for training and evaluation.

\section{Evaluation}

\subsection{Baseline Methods}

\subsubsection{Transformer}
We use the network design of BUDDI, \ie embedding, person, and parameter layers, the transformer encoder block and layers to bring the latents back into parameter space. The network takes BEV \cite{sun2022putting} estimates  as input and its task is to predict the correct \smplx parameters. We train this network on the same data as the conditional version of \mn. This baseline is equivalent to a single-shot (non-iterative) version of our diffusion model. 

\subsubsection{Contact Heuristic}
We design an optimization method which is similar to the routine we use to create Flickr Fits, but replaces the $L_{\contactmapbinary}$, \ie the loss that takes ground-truth contact maps into account, with a contact heuristic loss $L_{d_{\text{min}}}$. The contact heuristic loss encourages contact between the two people by minimizing their minimum distance. Given the vertices of each mesh, $v \in V_{X1}$ and $u \in V_{X2}$, we define the contact heuristic loss as
$$\energy_{d_{\text{min}}} = \min_{v,u} || v - u ||$$
and the overall objective function to be minimized becomes 
\begin{equation}
    \begin{split}
      L_{\text{Heuristic-fitting}} = & \lambda_{J} \energy_{J} + \lambda_{\bar{\theta}} \energy_{\bar{\theta}} +  \lambda_{\theta} \energy_{\theta} + \\
      & \lambda_{\shape} \energy_{\shape} + \lambda_{P} \energy_{P} + \lambda_{d_\text{min}} \energy_{d_{\text{min}}} \text{.}
    \end{split}
\end{equation}

\subsubsection{BUDDI (gen.)}

The conditional version of BUDDI can generate human meshes in close social interaction from noise given a BEV estimate. We use these generations to initialize the optimization routine and evaluate them against the ground truth. 

\subsubsection{VAE}

We also compare against VAE \cite{kingma2013auto} using the same training data. This model projects the \smplx parameters of two people into latent vectors of size 64, modeling a distribution, and from the latent space back into parameter space. Similar to the design of BUDDI, we embed each parameter via an MLP. We use two encoder and two decoder layers. The VAE training loss is 
$$L_{\text{VAE-training}} = L_{\theta} + L_{\beta} + L_{\gamma} + L_{v2v} +L_{\text{KL}} \text{.}$$

We use the same body model parameter losses as during BUDDI training. $L_{\text{KL}}$ is a standard KL-divergence loss between two Gaussians:
$$L_{\text{KL}} = \log{\frac{\sigma_2}{\sigma_1}} + \frac{\sigma^2_1 + (\mu_1 - \mu_2)^2}{2 \sigma^2_2} - \frac{1}{2}$$

During optimization, instead of optimizing body model parameters, we  optimize in the VAE's latent space. The optimization objective is:
\begin{equation}
    \begin{split}
L_{\text{VAE-fitting}} = & \lambda_{J} \energy_{J} + \lambda_{\bar{\theta}} \energy_{\bar{\theta}} +  \lambda_{\theta} \energy_{\theta} + \\
      & \lambda_{\shape} \energy_{\shape} + \lambda_{P} \energy_{P} + \lambda_{\text{VAE}} \energy_{\text{VAE}}\text{,}
    \end{split}
\end{equation}
where $\energy_{\text{VAE}}$ denotes a squared L2-loss on the VAE latent vector.

\subsection{Ablation of baseline methods}
We run our baseline methods under different conditions, \ie we use different weights for the Heuristic for a better comparison against the weights used in Flickr Fits and when optimizing with BUDDI used as a prior. The loss weights of Heuristic (a) are similar to those of Flickr Fits and the weights of Heuristic (b) to those of BUDDI. We report these numbers in \Cref{tab:hi4d_test_result_mpjpe_supp}, \Cref{tab:chi3d_test_result_mpjpe_supp}, and \Cref{tab:flickr_test_result_mpjpe_supp}.

\subsection{Perceptual study} 
We provide several quantitative evaluations of \mn in the main paper but there are aspects of human interaction that are subtle and best judged by people.
In the main part of this paper we present the results of the perceptual study that evaluates how realistic the generated interactions sampled from \mn are compared to meshes sampled from a VAE, the training data, and a random configuration of meshes.
Here, we show the layout and instructions of the perceptual study in \Cref{fig:mturk_user_study_layout}. We randomly sample 256 meshes from one training batch of size 512 created with a 60/20/20 ratio of meshes from Flickr/Hi4D/CHI3D. The meshes from the training batch are real samples from \mocap or by fitting \smplx to images with ground-truth contact map annotations. We sample 256 from BUDDI (unconditional model) and the VAE. To create the random baseline, we center all meshes in the training batch, shuffle the people along batch and person dimensions, and sample 256 mesh pairs. This is equivalent to real samples, except that each person are sampled randomly and not as a pair. 
Using Amazon Mechanical Turk (AMT), each participant was asked to rate 68 video comparisons per human intelligence task (HIT) with each video showing one pair of meshes at 360-degree views. Each HIT starts with 10 training videos (not used in evaluation) and contains 10 catch trials. Catch trials show implausible interaction, \eg two people with random poses placed on top of each other. The training videos are presented at the beginning of the task, and the method and catch trial videos appear in random order. The remaining 48 comparisons show one sample from BUDDI against either VAE / random baseline / or training data (12 comparisons per method). We randomly shuffle the video order per HIT and left / right. Each HIT is conducted by 6 participants. We exclude HITS where participants fail three or more catch trials. Our final results were computed with the responses from the 83/96 participants who passed.

\subsection{Additional qualitative results and failure cases} 
We provide additional qualitative examples of optimization with \mn and compare them to optimization with heuristics and \bev in \Cref{fig:sup_qual_01} and \Cref{fig:sup_qual_02}. Failure cases are provided in \Cref{fig:sup_failure}.


\input{sections/03_BUDDI/tables/supmat_tables}

\input{sections/03_BUDDI/figtex/supmat}

\clearpage 
\input{sections/03_BUDDI/tables/supmat_code}
\clearpage

%% file: sections/03_BUDDI/figtex/psgt_supmat.tex
\begin{figure}
    \centering
\includegraphics[width=\linewidth,clip,trim=0cm 8cm 0cm 32cm]{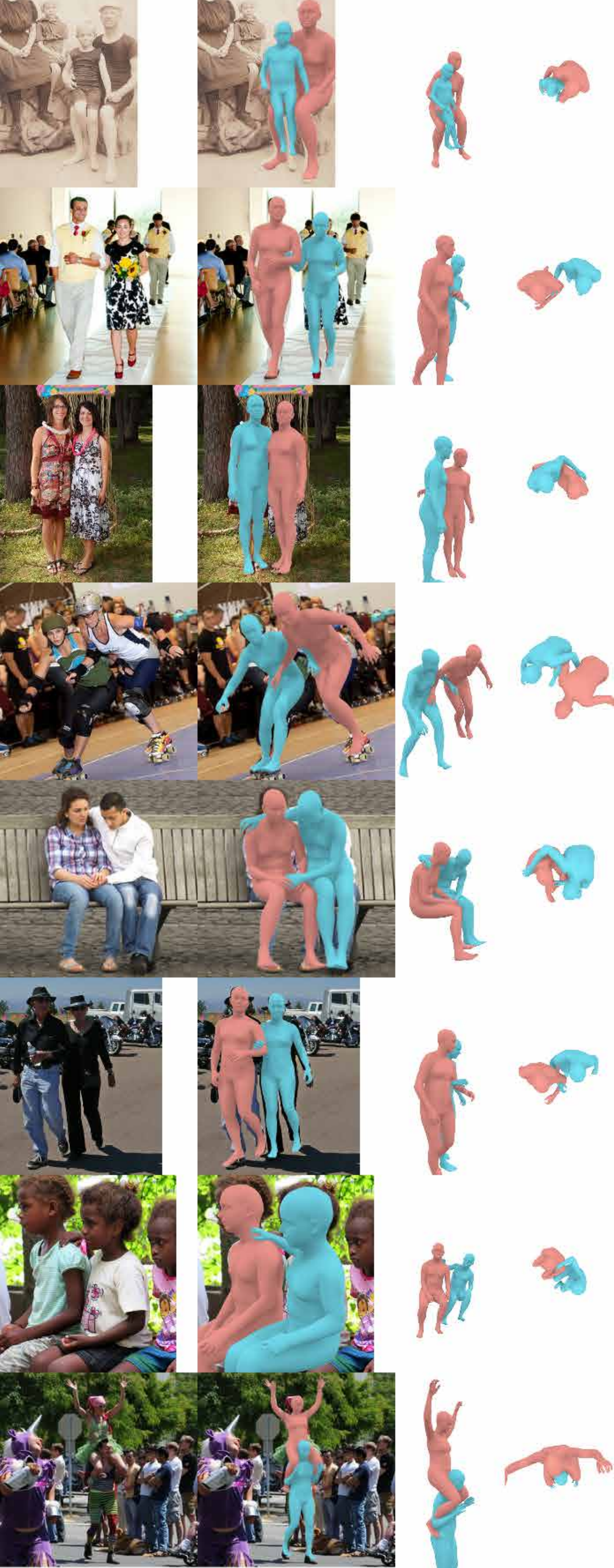}
\caption[Flickr Fits]{\textbf{Flickr Fits.} We visualize the output of the optimization process that reconstructs two people in close proximity using ground-truth contact maps, shown from three different views.} \vspace{-1em}
\label{fig:psgt_supmat}
\end{figure}

%% file: sections/03_BUDDI/tables/supmat_tables.tex
\begin{table*}
\scriptsize
\begin{center}
        \begin{tabular}{llllllllllll}
        \toprule[0.4mm]
                                       & $\lambda_{J2D}$     & $\lambda_{\tilde{\theta}}$ & $\lambda_{\theta}$ & $\lambda_{\beta}$ & $\lambda_{C^B}$ & $\lambda_{d_{\text{min}}}$ & $\lambda_{P}$         & $\lambda_{\theta_{\text{\mn}}}$ & $\lambda_{\trans_{\text{\mn}}}$ & $\lambda_{\beta_{\text{\mn}}}$ & $\lambda_{\text{VAE}}$ \\ 
        \midrule

        Flickr Fits    & 0.04/0.1  & 200/200 & 4/4   & 40/0     & 10/10  & 0/0     & 0/1000 & 0/0     & 0/0     & 0/0     & 0/0 \\
        \mn            & 0.02/0.02 & 200/200 & 0/0   &  0/0     & 0/0    & 0/0     & 0/10   & 100/100 & 10/10   & 1e5/1e5 & 0/0 \\
        VAE            & 0.02/0.1  & 200/200 & 2/2   & 40/0     & 0/0    & 0/0     & 0/0.1  & 0/0     & 0/0     & 0/0     & 1/1 \\
        Heuristics     & 0.02/0.1  & 200/200 & 2/2   & 40/0     & 0/0    & 1e5/1e5 & 0/0.1  & 0/0     & 0/0     & 0/0     & 0/0 \\
        Heuristics (a) & 0.04/0.1  & 200/200 & 4/4   & 40/0     & 0/0    & 1e5/1e5 & 0/1000 & 0/0     & 0/0     & 0/0     & 0/0 \\
        Heuristics (b) & 0.02/0.02 & 200/200 & 4/4   & 40/0     & 0/0    & 1e5/1e5 & 0/10   & 0/0     & 0/0     & 0/0     & 0/0 \\

        \bottomrule[0.4mm]
        \end{tabular}
\end{center}
\caption[Weights of the different loss terms in optimization]{\textbf{Weights of the different loss term during the optimization.} We consider the case of using \pgt contact maps, the heuristics, and \mn. Optimizations with \mn and \pgt are run for two stages. The optimization with heuristics converges quickly so a single stage is enough.}
\label{tab:training_weights}
\end{table*}

\begin{table*}
\scriptsize 
    \centering
        \begin{tabular}{lcc|cccccccccccc} 
        \toprule[0.4mm]
     &    PER PERSON $\downarrow$ & JOINT $\downarrow$ & \multicolumn{11}{c}{JOINT PA-MPJPE $\downarrow$} \\
     & PA-MPJPE & PA-MPJPE & backhug    & basketball & cheers     & dance      & fight      & highfive   & hug        & kiss       & pose       & sidehug    & talk                   \\
    \midrule
    Heuristic     & 67  / 71        & 121        & 168        & 83         & 94         & 131        & 94         & 68         & 159        & 159        & 118        & 113        & 109       \\
    Heuristic (a) & 68  / 72        & 122        & 166        & 82         & 93         & 126        & 92         & 68         & 161        & 158        & 122        & 122        & 114       \\
    Heuristic (b) & 68  / 73        & 124        & 164        & 90         & 92         & 130        & 95         & 68         & 161        & 158        & 125        & 124        & 117       \\
        \bottomrule[0.4mm]
        \end{tabular}
    \caption[Evaluation of BUDDI on Hi4D]{\textbf{Evaluation of BUDDI on Hi4D.} We compare the output of \mn to the proposed baseline methods on the Hi4D challenge. The first block shows methods that do not use \hifourd data during training or are optimization based without access to priors trained on \hifourd. \mn (F,C) in particular, is our model \mn trained on \flickr and CHI3D data only. All errors are reported in mm for 3D Joints.}
    \label{tab:hi4d_test_result_mpjpe_supp}
\end{table*} 

\begin{table}
\centering
\resizebox{.7\columnwidth}{!}{

    \begin{tabular}{lccc}
        \toprule[0.4mm]
                    & PER PERSON  $\downarrow$           & JOINT $\downarrow$   \\
                    & PA-MPJPE               & PA-MPJPE \\
        \midrule
    Heuristic       & 49     \ 46     & 105     \\
    Heuristic  (a)  & 49     \ 47     & 103     \\
    Heuristic  (b)  & 47     \ 45     & 103     \\
        \bottomrule[0.4mm]
    \end{tabular}}
    \caption[Ablation study of baseline methods on CHI3D]{\textbf{Quantitative Evaluation on \chithreed.} We compare different versions of the baseline optimization with contact heuristic on \chithreed (pair s03). All errors reported in mm for 3D Joints.}
    \label{tab:chi3d_test_result_mpjpe_supp}
\end{table}  

\addtolength{\tabcolsep}{-2pt}
\begin{table}
\scriptsize
\begin{center}
\begin{tabular}{lcccccccccccccc}
        \toprule[0.4mm]
        &    \multirow{2}{*}{\shortstack{JOINT $\downarrow$ \\ PA-MPJPE}}  &  \multicolumn{5}{c}{PCC at radius $\uparrow$}  \\
        &       &   5 & 10 & 15 & 20 & 25 \\ 
        \midrule
Heuristic      & 68      & 14      & 34      & 49      & 61      & 70       \\
Heuristic (a)  & 69      & 11      & 30      & 45      & 57      & 66       \\
Heuristic (b)  & 72      & 12      & 30      & 45      & 57      & 67       \\
        \bottomrule[0.4mm]
\end{tabular}
\end{center}
\vspace{-0.8em}
\caption[Ablation study of baseline methods on \flickrcithreeds]{\textbf{3D Pose Evaluation on \flickrcithreeds.} We compare different versions of the baseline optimization with contact heuristic on the \flickr fits using their joint (two-person) PA-MPJPE expressed in mm. We also evaluate the percentage of correct contact points (PCC) for radius \textit{r} mm.
}
\label{tab:flickr_test_result_mpjpe_supp}
\end{table}
\addtolength{\tabcolsep}{3pt}

%% file: sections/03_BUDDI/figtex/supmat.tex
\begin{figure*}
\includegraphics[width=0.9\textwidth]{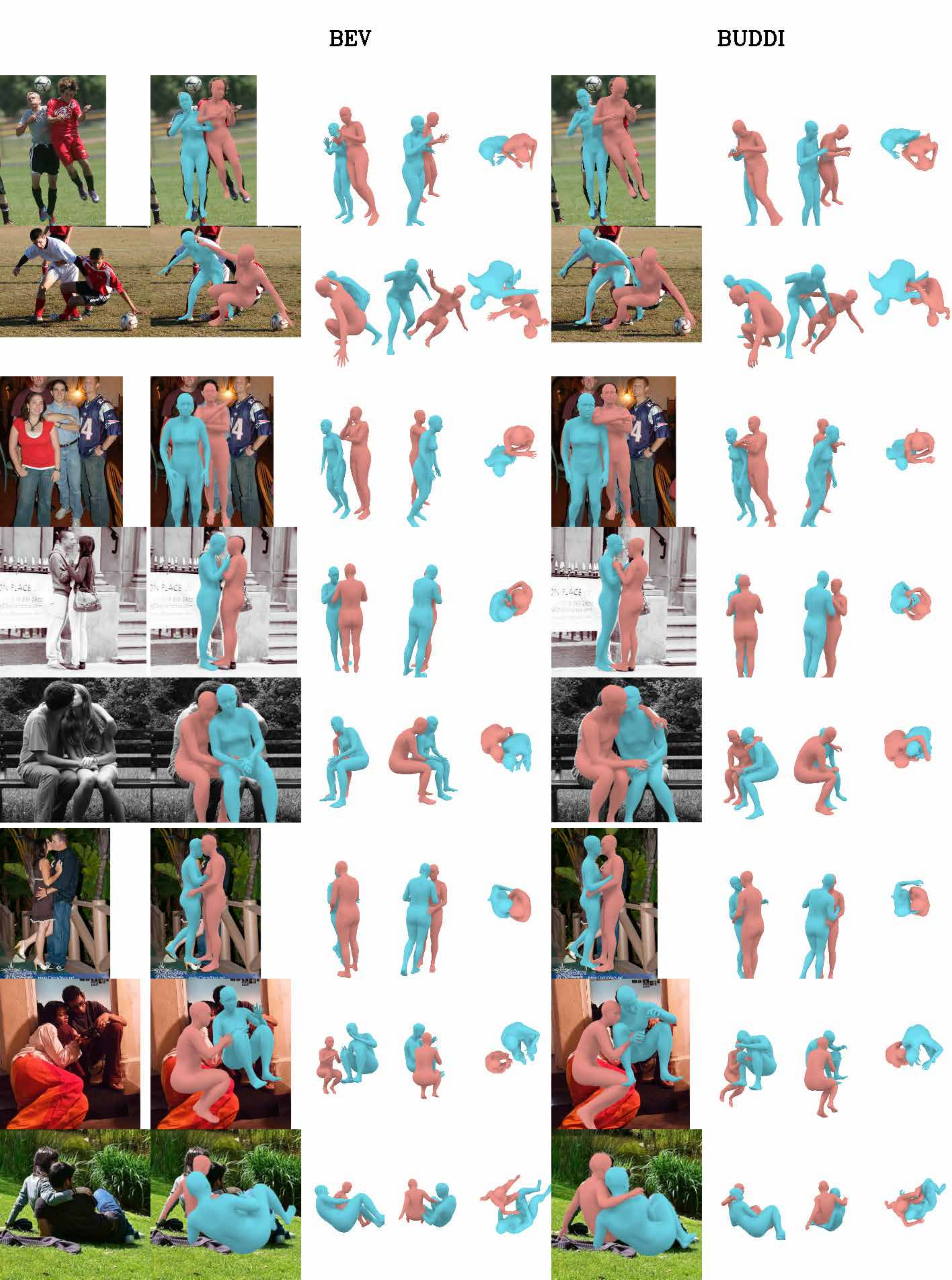}
\centering
\caption[Qualitative examples from optimization with \mn]{\textbf{Optimization with \mn.} Additional qualitative examples from optimization with \mn compared to \bev. We provide the overlay and three additional views per method. Optimization with \bev (first method / columns 2-5), optimization with \mn (second method / columns 6-9).}
\label{fig:sup_qual_01}
\end{figure*}

\begin{figure*}
\includegraphics[width=0.9\textwidth]{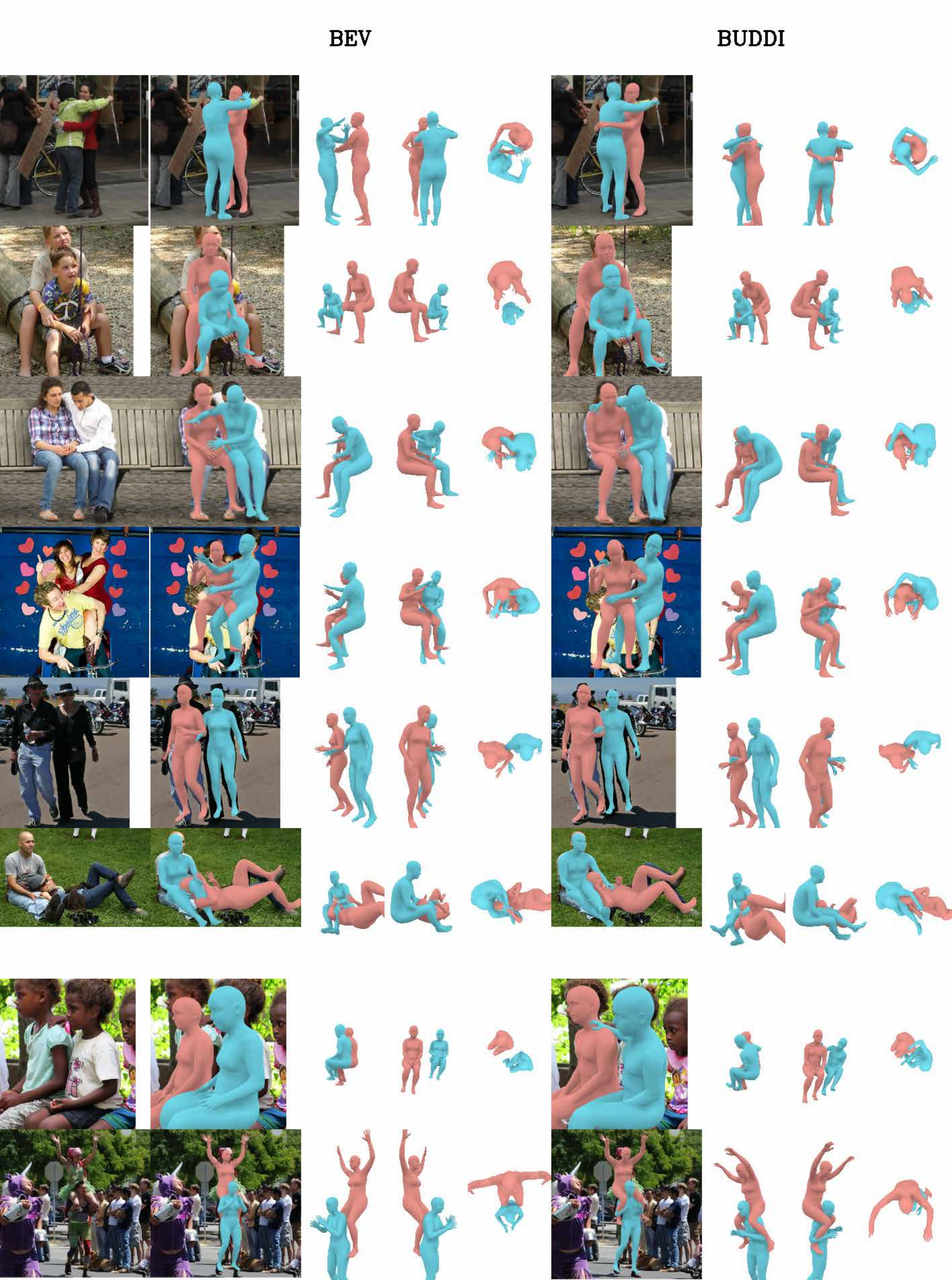}
\centering
\caption[Qualitative examples from optimization with \mn]{\textbf{Optimization with \mn (continuation).} Additional qualitative examples from optimization with \mn compared to \bev. We provide the overlay and three additional views per method. Optimization with \bev (first method / columns 2-5), optimization with \mn (second method / columns 6-9).}
\label{fig:sup_qual_02}
\end{figure*}

\begin{sidewaysfigure*}
\includegraphics[trim={0 10cm 0 0},clip,width=\textwidth]{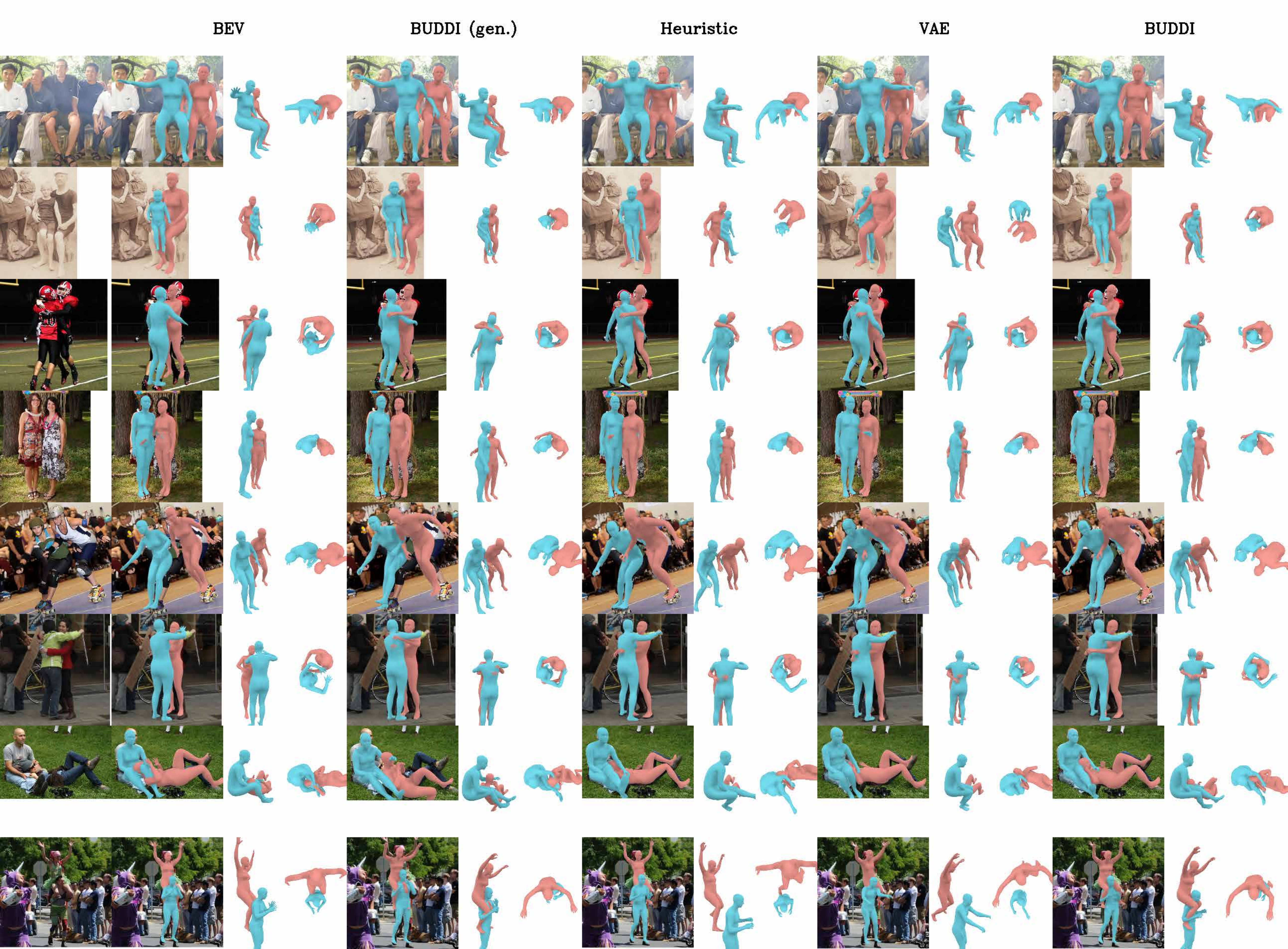}
\centering
\caption[Qualitative examples from optimization with \mn]{\textbf{Optimization with \mn.} Additional qualitative examples from optimization with \mn compared to \bev, BUDDI generations, optimization with heuristic, and optimization with VAE. We provide the overlay and three additional views per method. \bev (first method / columns 2-5), \mn (gen.) (second method / columns 6-9), optimization with heuristic (third method / columns 10-13), optimization with VAE (fourth method / columns 14-16), and optimization with \mn (fifth method / columns 17-20) .}
\label{fig:sup_qual_02_all}
\end{sidewaysfigure*}

\begin{figure*}
\includegraphics[width=0.75\linewidth]{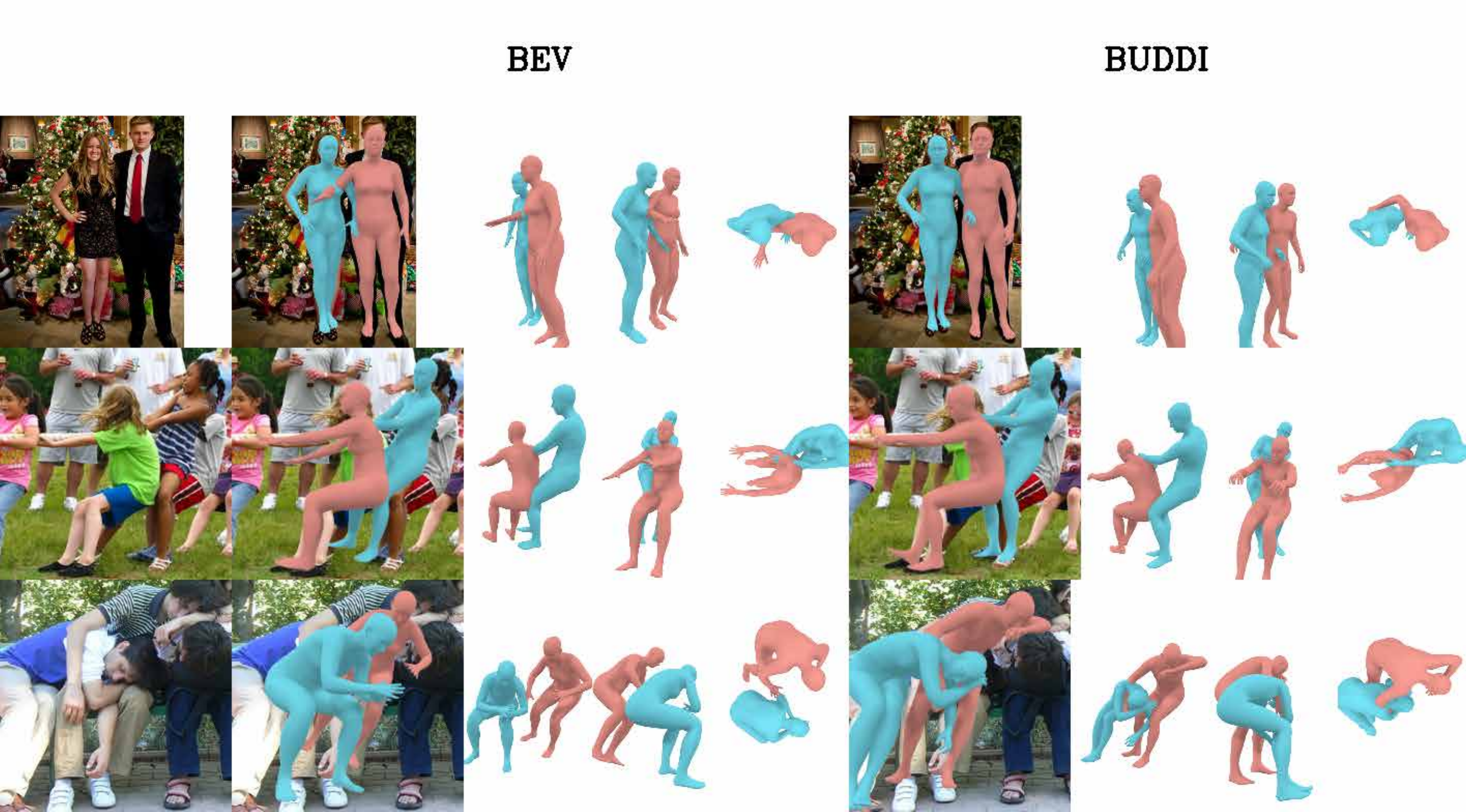}
\centering
\caption[Failure cases optimization with \mn]{\textbf{Failure cases optimization with \mn.} Failure cases from optimization with \mn. In the first row the depth ordering of leg arm is wrong. The image in row~ 2 contains less common contact so that \mn suggests for blue to hold red's shoulders instead of the rope. The estimated predicted by our method suggests a plausible pose that is not consistent with the image due to single-view ambiguity. The last row shows a failure case due to intersection between arm and torso.}
\label{fig:sup_failure}
\end{figure*}

\begin{figure*}
\includegraphics[width=0.8\textwidth]{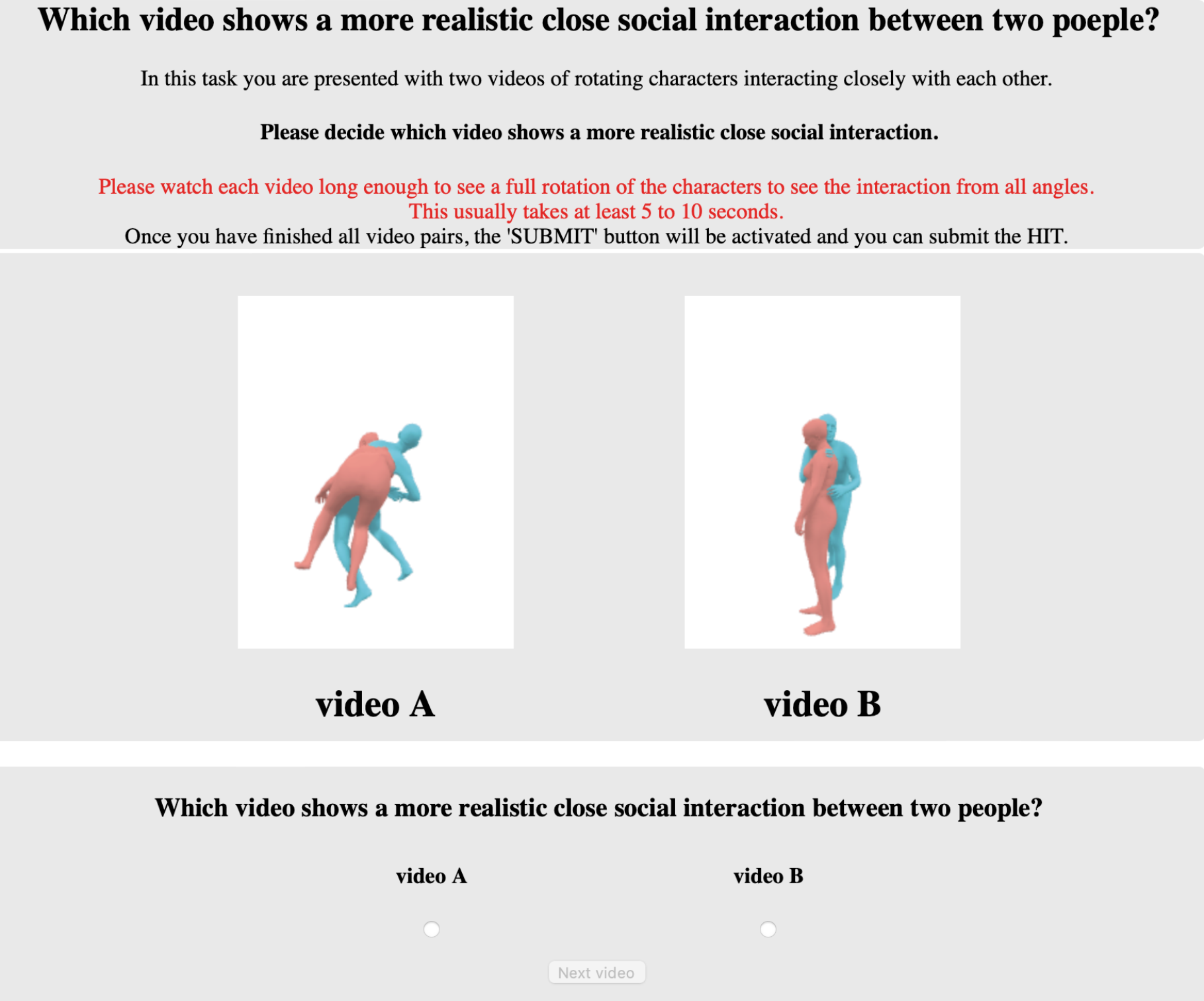}
\centering
\caption[Amazon Mechanical Turk perceptual study layout and instructions]{{\textbf{Amazon Mechanical Turk perceptual study layout and instructions.}} On the left, we show a 360-degree video of the two interacting people. On the right, the rating scale.}
\label{fig:mturk_user_study_layout}
\end{figure*}

%% file: sections/03_BUDDI/tables/supmat_code.tex
\clearpage
\begin{lstlisting}[language=Python, caption=Pseudo code for optimization with BUDDI., label=buddi_opti]
import smplx
import buddi

# optimization params
num_stages = 2
max_iterations = 100 
t = 10 # noise level

# create smpl and buddi
smpl = smplx.create(model_folder)

# load buddi denoiser model (D)
buddi = buddi.create(checkpoint_path).eval()

# load detected keypoints and bev
kpts = load_keypoint_detections(img_path)
bev = load_bev_estimate(img_path)

# sample from buddi conditioned on BEV
buddi_sample = sample_from_buddi(cond=bev)

# initialize the optimization
smpl.params = buddi_sample

# run optimization
for ss in range(num_stages):
  optimizer = setup_optimizer(smpl, ss)
    
  for ii in range(max_iterations):
    # fitting losses
    fitting_loss = get_fitting_loss(
                        smpl, buddi_sample, kpts)
    
    # detach current smpl, then diffuse & denoise
    with torch.no_grad():
      diffused_smpl = smpl + sample_noise(t)
      denoised_smpl = buddi(diffused_smpl, t)
      
    # compute diffusion losses
    diffusion_loss = get_diffusion_loss(
                        smpl, denoised_smpl)      
    
    # final loss of iteration ii of stage ss
    total_loss = fitting_loss + diffusion_loss
    
    # backprop
    optimizer.zero_grad()
    total_loss.backward()
    optimizer.step()

    # check stopping criterium
    if converted:
      break

\end{lstlisting}
\clearpage